%% file: main.tex
\documentclass[a4paper,fleqn]{cas-dc}
\usepackage[numbers, compress]{natbib}

\usepackage{subcaption} 
\usepackage{caption}
\usepackage{xurl} 

\usepackage{amsmath,amsfonts}
\usepackage{algorithmic}
\usepackage{array}

\usepackage{textcomp}
\usepackage{stfloats}
\usepackage{float}
\usepackage{url}
\usepackage{verbatim}
\usepackage{graphicx}

\def\BibTeX{{\rm B\kern-.05em{\sc i\kern-.025em b}\kern-.08em
T\kern-.1667em\lower.7ex\hbox{E}\kern-.125emX}}
\usepackage{balance}

\usepackage{booktabs, makecell, multirow, tabularx}
\usepackage{microtype}    
\usepackage{hyperref}
\hypersetup{
  colorlinks=true,
  linkcolor=blue,
  citecolor=green,
  urlcolor=magenta,
  filecolor=cyan,
}

\usepackage{cleveref}     
\usepackage{wrapfig} 
\usepackage[dvipsnames,table]{xcolor} 
\crefname{figure}{Fig.}{Figs.}
\crefname{table}{Table}{Tables}
\crefname{equation}{Eq.}{Eqs.}
\everymath{\thickmuskip=2mu minus 2mu} 

\usepackage{amssymb}
\usepackage{pifont}

\usepackage{enumitem}

\def\tsc#1{\csdef{#1}{\textsc{\lowercase{#1}}\xspace}}
\tsc{WGM}
\tsc{QE}

\begin{document}
\let\WriteBookmarks\relax
\def\floatpagepagefraction{1}
\def\textpagefraction{.001}
\let\printorcid\relax 

\shorttitle{Adapting Vision-Language Foundation Model for Next Generation Medical Ultrasound Image Analysis}

\shortauthors{<short author list for running head>}
\shortauthors{Jingguo Qu et al.}

\title[mode = title]{Adapting Vision-Language Foundation Model for Next Generation Medical Ultrasound Image Analysis}

\author[1]{Jingguo Qu}[style=chinese]
\cormark[1]
\credit{Conceptualization, Investigation, Writing original draft, Writing review \& editing}

\author[1]{Xinyang Han}[style=chinese]
\cormark[1]
\credit{Investigation, Writing - review \& editing}

\author[4]{Jia Ai}[style=chinese]
\credit{Data curation}

\author[4]{Juan Wu}[style=chinese]
\credit{Data curation}

\author[5]{Tong Zhao}[style=chinese]
\credit{Data curation}

\author[1]{Tonghuan Xiao}[style=chinese]
\credit{Data curation}

\author[1]{Sheng Ning}[style=chinese]
\credit{Data curation}

\author[1]{Yuqi Yang}[style=chinese]
\credit{Data curation}

\author[2]{Jing Qin}[style=chinese]
\credit{Supervision}

\author[3]{Ann Dorothy King}[style=chinese]
\credit{Supervision}

\author[3]{Winnie Chiu-Wing Chu}[style=chinese]
\credit{Supervision}

\author[1]{Jing Cai}[style=chinese]
\credit{Supervision}

\author[1]{Michael Tin-Cheung Ying}[style=chinese]
\cormark[2]
\ead{michael.ying@polyu.edu.hk}
\credit{Conceptualization, Supervision, Writing - review \& editing}

\affiliation[1]{organization={Department of Health Technology and Informatics, The Hong Kong Polytechnic University},
	city={Hong Kong},
	country={China}
}
\affiliation[2]{organization={Centre for Smart Health and School of Nursing, The Hong Kong Polytechnic University},
	city={Hong Kong},
	country={China}
}
\affiliation[3]{organization={Department of Imaging and Interventional Radiology, The Chinese University of Hong Kong},
	city={Hong Kong},
	country={China}
}
\affiliation[4]{organization={Suzhou Hospital of Traditional Chinese Medicine Affiliated to Nanjing University of Chinese Medicine},
	city={Suzhou},
	country={China}
}
\affiliation[5]{organization={Department of Ultrasound, The Affiliated Changzhou No. 2 People's Hospital of Nanjing Medical University},
	city={Changzhou},
	country={China}
}

\cortext[1]{Equal contribution}
\cortext[2]{Corresponding author}

\input{sec/0_abstract}


\maketitle

\input{sec/1_intro}
\input{sec/2_related_work}
\input{sec/3_method}
\input{sec/4_setup}
\input{sec/5_results}
\input{sec/6_conclusion}
\input{sec/7_credits}

\bibliographystyle{cas-model2-names}
\bibliography{ref}

\clearpage
\twocolumn[
	\begin{center}
		\Large\textbf{Supplementary Material}
	\end{center}
	\vspace{1em}
]
\input{sec/8_appendix}

\end{document}

%% file: sec/0_abstract.tex
\begin{abstract}
	Vision-Language Foundation Models (VLFMs) exhibit remarkable generalization, yet their direct application to medical ultrasound is severely hindered by a profound modality gap. The unique acoustic physics of ultrasound, characterized by speckle noise, shadowing, and heterogeneous textures, often degrades the performance of off-the-shelf VLFMs. To bridge this gap, we propose a novel Hybrid Tuning (HT) strategy for the parameter-efficient adaptation of CLIP-based models to ultrasound analysis. Instead of updating the pre-trained weights, HT freezes the visual backbone and integrates a specialized lightweight adapter. This adapter features a Frequency Filtering module to suppress domain-specific periodic artifacts and a Noise Estimation module to dynamically calibrate feature representations. Extensive evaluations across six multi-center datasets demonstrate that our HT-enhanced models significantly outperform existing state-of-the-art adapters and medical VLFMs in both segmentation and classification tasks. Notably, HT exhibits exceptional data efficiency in few-shot scenarios and robust cross-dataset generalization. Our findings prove that preserving pre-trained semantic priors while explicitly modeling ultrasound-specific noise is key to unlocking foundational intelligence in automated ultrasound diagnosis. The source code is available at \href{https://github.com/jinggqu/NextGen-UIA}{GitHub}.
\end{abstract}

%% file: sec/1_intro.tex
\section{Introduction}

Ultrasonography is a common medical imaging technique used in clinical practice. It is particularly favored for evaluating superficial structures, including breast lesions~\cite{screening_brem_2015}, thyroid nodules~\cite{thyroid_takashima_1995}, and lymph nodes~\cite{sonography_ying_2003,sonographic_ahuja_2005,ultrasound_ahuja_2008}, due to its ability to provide real-time visualization in a noninvasive and cost-effective manner. Utilizing high-frequency ultrasound, ultrasonography delivers detailed views of internal anatomy and is routinely used in clinical examinations. Currently, radiologists need to manually identify and delineate regions of interest (ROIs) on ultrasound images to diagnose diseases and develop treatment plans, which is a time-consuming process and subject to variability depending on the experiences of radiologists.

However, the inherent imaging vulnerabilities of ultrasonography pose challenges to this process, including grainy appearance, ill-defined ROI boundaries, low contrast, and the presence of image artifacts. Furthermore, the shapes, sizes, and locations of ROIs can vary substantially across scan planes and patient anatomies. These factors make manual annotation laborious and inconsistent, underscoring the need for automated methods that can reliably detect, segment and classify ROIs to improve diagnostic efficiency and accuracy. Thus, automated ultrasound image analysis is a potential direction for standardized and regulated diagnosis and treatment.

Deep learning has driven significant advances in medical image analysis~\cite{ai_han_2025,diff_han_2024}, but these methods rely on large volumes of expertly annotated data~\cite{unet_ronneberger_2015} and lack generalizability~\cite{ln_qu_2025}. Foundation models (FMs)~\cite{foundation_models_bommasani_2021}, which are pre-trained on vast and diverse datasets, offer a way to relax this requirement. By capturing broad visual representations during pre-training, they can be efficiently fine-tuned with few labeled ultrasound scans, while still maintaining coherent, task-relevant features. This paradigm holds great promise for more practical, data-efficient analysis in clinical practice. Yet, despite growing interest in adapting FMs to specialized domains, their application to ultrasound image analysis remains under-explored.

Mainstream FMs (\eg, CLIP~\cite{clip_radford_2021}, Segment Anything Model (SAM)~\cite{sam_kirillov_2023}, DINOv2~\cite{dinov2_oquab_2023}) have shown impressive performance in various computer vision tasks, such as zero-shot classification~\cite{clip_radford_2021}, semantic segmentation~\cite{sam_kirillov_2023,dinov2_oquab_2023} and visual-language interactions~\cite{omnivl_wang_2022}. Recently, ultrasound-specific FMs have emerged to tackle domain-specific challenges. For instance, USFM~\cite{usfm_jiao_2024} integrates multi-organ data for generalizable analysis, while UltraSAM~\cite{ultrasam_meyer_2024} adapts the SAM for ultrasound segmentation using limited data. However, training such specialized FMs from scratch or conducting full-model fine-tuning requires massive domain-specific datasets and computational resources. Moreover, as noted by Poudel~\etal~\cite{exploring_poudel_2023}, models trained solely on smaller medical datasets may lack the robust generalizability inherent in original FMs pre-trained on extensive natural image datasets. Therefore, developing efficient domain adaptation techniques that can bridge the gap between powerful, general-purpose FMs and specific ultrasound tasks without requiring extensive retraining remains critical and promising.

To address this problem, we propose a novel Parameter-Efficient Fine-Tuning (PEFT) framework Hybrid Tuning (HT) to adapt CLIP-based models~\cite{clip_radford_2021,biomedclip_zhang_2023,unimedclip_khattak_2024,medclip_wang_2022} for medical ultrasound image analysis, which is designed to be seamlessly integrated into the vision transformer backbone of CLIP. Our methodology involves fine-tuning the pre-trained CLIP and its variants on a radiological image-text dataset, where only the lightweight HT adapter is trained. The resulting adapted model is then evaluated on a comprehensive suite of downstream tasks, including zero-shot classification, supervised classification, and semantic segmentation on various ultrasound datasets. These datasets include the in-house lymph node ultrasound datasets from two different centers (LN-INT, LN-EXT), the publicly available breast ultrasound image (BUSI) dataset~\cite{busi_al_2020}, thyroid datasets DDTI~\cite{ddti_pedraza_2015} and TN3K~\cite{tn3k_gong_2023}, and microultrasound scans of prostate~\cite{microsegnet_jiang_2023}. Our experimental results demonstrate that this HT technique significantly outperforms baseline methods, establishing a new SOTA for adapting VLFMs to the medical ultrasound domain. The main contributions of this work are summarized as follows:
\begin{itemize}[leftmargin=*]
	\item We propose HT, a novel PEFT adapter specifically designed to bridge the domain gap between natural and ultrasound images for CLIP-based FMs.
	\item We introduce a Frequency Filtering module to suppress periodic ultrasound artifacts in the frequency domain, and a Noise Estimation module to dynamically calibrate features based on local speckle noise characteristics.
	\item We conduct extensive experiments on six datasets across four anatomical regions (lymph node, breast, thyroid, and prostate). The results demonstrate that HT consistently outperforms SOTA PEFT methods in both classification and semantic segmentation tasks, exhibiting superior cross-domain adaptability.
\end{itemize}

%% file: sec/2_related_work.tex
\section{Related Work}

\subsection{VLFMs in Medicine}

FMs~\cite{foundation_models_bommasani_2021}, pre-trained on vast datasets via self-supervised learning, have demonstrated remarkable transferability across tasks. Prominent examples include GPT~\cite{gpt_radford_2019}, BERT~\cite{bert_devlin_2018}, SAM~\cite{sam_kirillov_2023}, and DINOv2~\cite{dinov2_oquab_2023}. Among these, CLIP~\cite{clip_radford_2021} introduces a contrastive framework that aligns visual and textual representations, enabling zero-shot classification without task-specific training.

In the medical domain, researchers have explored two primary strategies to leverage CLIP. The first involves adapting the original CLIP model to medical tasks. For instance, PubMedCLIP~\cite{pubmedclip_eslami_2021} and MedCLIP~\cite{medclip_wang_2022} fine-tune CLIP for visual question answering (VQA), while CLIPSeg~\cite{clipseg_luddecke_2022} extends it to dense prediction tasks. The second strategy focuses on training CLIP-style models from scratch using large-scale biomedical data. PMC-CLIP~\cite{pmcclip_lin_2023} utilized 1M image-text pairs from PubMed Central to achieve SOTA results in retrieval and VQA. BiomedCLIP~\cite{biomedclip_zhang_2023} scaled this to 15M pairs, and UniMedCLIP~\cite{unimedclip_khattak_2024} further introduced a unified model trained on 5.3M open-source samples across diverse modalities.

Despite these advancements, significant challenges remain. First, even the largest biomedical datasets (\eg, 15M in BiomedCLIP) are dwarfed by the generic CLIP dataset (400M), limiting the robustness and generalization, as discussed in MetaCLIP~\cite{metaclip_xu_2023} and Poudel~\etal~\cite{exploring_poudel_2023}. Second, most medical VLFMs focus on sparse prediction tasks (\eg, classification), with dense tasks (\eg, segmentation) often under-explored. This necessitates efficient domain adaptation methods that can leverage the strong priors of general FMs while addressing the specific characteristics of medical imaging.

\subsection{Ultrasound Foundation Models}

Recent efforts have emerged to develop FMs specifically tailored for ultrasound. Jiao~\etal introduced USFM~\cite{usfm_jiao_2024}, a universal ultrasound model trained on a large-scale multi-organ dataset to improve cross-task generalization. Similarly, Meyer~\etal proposed UltraSAM~\cite{ultrasam_meyer_2024}, adapting the SAM for robust ultrasound segmentation with limited supervision. While effective, these domain-specific vision-only models typically require training from scratch or extensive fine-tuning on large, curated ultrasound datasets, which are often difficult to obtain due to privacy concerns and annotation costs. This highlights a critical need for methods that can efficiently adapt existing powerful FMs to the ultrasound domain without relying on massive domain-specific training data.

\subsection{Parameter-Efficient Fine-Tuning}

Adapting large-scale FMs to specific downstream tasks presents a significant challenge: full fine-tuning is often computationally prohibitive and prone to overfitting, particularly in medical scenarios with limited annotated data. PEFT has emerged as an effective solution, freezing the pre-trained backbone while optimizing a minimal set of extra parameters. Originating in the NLP field, methods like Adapter~\cite{peft_houlsby_2019} insert lightweight modules into transformer layers, while LoRA~\cite{lora_hu_2022} injects learnable low-rank matrices into attention blocks to approximate weight updates.

These techniques have been successfully translated to the vision domain. VPT~\cite{vpt_jia_2022} introduces learnable prompts to the input sequence, and AdaptFormer~\cite{adaptformer_chen_2022} employs parallel bottleneck adapters to enhance Vision Transformer (ViT) for classification. While effective for recognition tasks, these standard approaches often struggle with dense prediction tasks, which require fine-grained spatial information. Although recent works like Mona~\cite{mona_yin_2024} have attempted to bridge this gap using multi-scale adapters, they remain rooted in general visual feature adaptation and overlook the specific domain shift of ultrasound imaging. Consequently, they may fail to capture the subtle boundary details required for precision medical diagnosis. Our HT module addresses this limitation by integrating frequency-domain filtering with noise-adaptive spatial attention, ensuring robust adaptation to the ultrasound domain without compromising parameter efficiency.

%% file: sec/3_method.tex
\section{Method}

\subsection{Overview}

\begin{figure*}[pos=htb]
	\centering
	\includegraphics[width=0.8\linewidth]{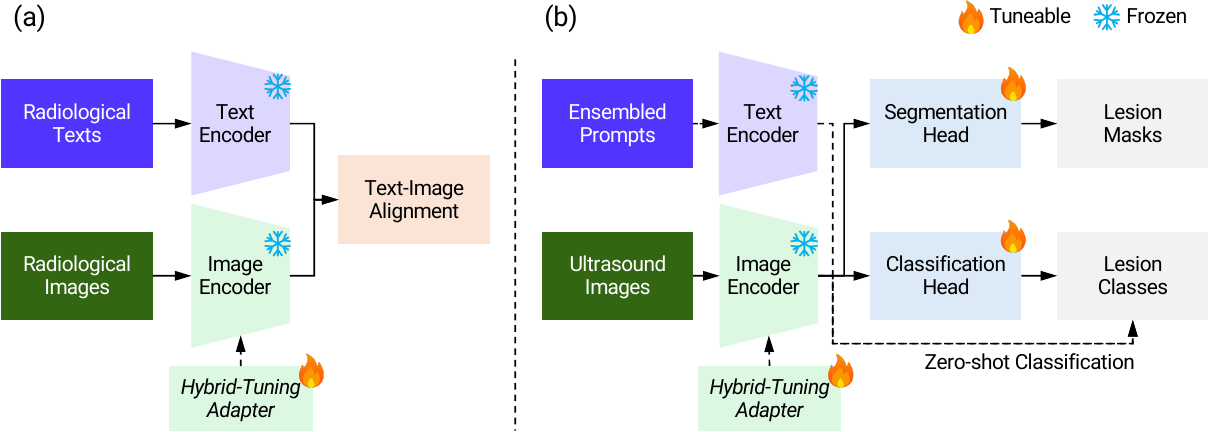}
	\caption{Overview of proposed workflow. (a) \textbf{Fine-tuning stage}. Introduce trainable HT adapter into frozen CLIP to bridge the domain gap between natural images and radiological scans. (b) \textbf{Downstream tasks}. Apply trainable heads for ultrasound image segmentation and classification in a supervised manner (solid arrows), and assess zero-shot ultrasound diagnosis capability of CLIP by using ensembled prompt-image pairs (dashed arrows).}
	\label{fig:overview}
\end{figure*}

As illustrated in~\cref{fig:overview}, our approach inserts lightweight HT adapters into the visual encoder layers. These adapters are explicitly designed to disentangle and suppress ultrasound-specific artifacts in the frequency domain while dynamically calibrating features based on noise levels. This allows the model to learn robust, clearer representations without altering the pre-trained weights. Finally, task-specific heads are attached to the adapted encoder for segmentation and classification.

\subsection{Hybrid-tuning Adapter}

\begin{figure}[pos=htb]
	\centering
	\includegraphics[width=\linewidth]{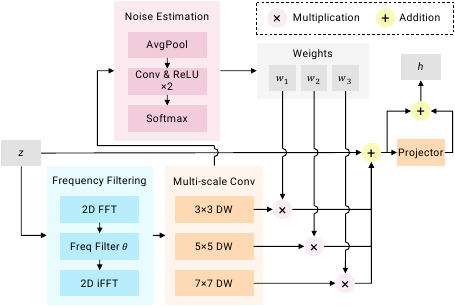}
	\caption{Architecture of Hybrid-tuning module.}
	\label{fig:ht}
\end{figure}

While inspired by the architecture of Mona~\cite{mona_yin_2024}, HT is tailored for the standard ViT~\cite{vit_dosovitskiy_2020} backbone and introduces specific mechanisms for ultrasound artifact suppression by appending after every transformer block in the vision encoder. As shown in~\cref{fig:ht}, HT consists of two key components: a Frequency Filtering (FF) module for periodic artifact removal and a Noise Estimation (NE) module for dynamic feature calibration.

Formally, let $\mathbf{Z} \in \mathbb{R}^{N \times B \times D}$ denote the input token sequence. Following the implementation, we first apply normalized residual scaling and then project to a bottleneck dimension $d$:
\begin{equation}
	\mathbf{Z}_{in} = (\text{LayerNorm}(\mathbf{Z}) \odot \gamma + \mathbf{Z} \odot \gamma_x) \mathbf{W}_{down} + \mathbf{b}_{down}
\end{equation}
where $\mathbf{W}_{down} \in \mathbb{R}^{D \times d}$ is the projection matrix. $\mathbf{Z}_{in}$ is reshaped into a 2D feature map $\mathbf{F}_{in} \in \mathbb{R}^{B \times d \times H \times W}$.

\textbf{FF Module.}
Ultrasound images often contain periodic artifacts (such as reverberations, side lobes, and scan-line artifacts) arising from the physics of sound wave propagation. In the spatial domain, these global artifacts are entangled with tissue structures. However, in the frequency domain, they manifest as distinct high-energy peaks. To suppress these, we employ a learnable channel-wise filter $\boldsymbol{\theta} \in \mathbb{R}^d$, applied via a real-valued 2D FFT:
\begin{equation}
	\mathbf{F}_{freq} = \text{irFFT2}\left(\text{rFFT2}(\mathbf{F}_{in}) \odot \boldsymbol{\theta}\right)
\end{equation}
where $\boldsymbol{\theta}$ is broadcast along spatial-frequency dimensions, and $\mathbf{F}_{freq} \in \mathbb{R}^{B \times d \times H \times W}$ is the frequency-filtered feature map returned to the spatial domain. Our spectral analysis confirms that trained $\boldsymbol{\theta}$ significantly attenuates spectral energy in artifact-related bands, validating its physical role as a flexible band-stop filter.

\textbf{NE Module.}
Speckle noise in ultrasound is granular and signal-dependent, requiring adaptive smoothing. Fixed-kernel filtering risks blurring anatomical details. The NE module dynamically estimates the noise level of the input features and adjusts the receptive field accordingly.
Given the frequency-filtered feature map $\mathbf{F}_{freq}$, we compute channel-wise mixture weights $\mathbf{w} \in \mathbb{R}^{B \times 3 \times 1 \times 1}$ using global average pooling (GAP), two $1\times 1$ convolutions, and softmax normalization:
\begin{equation}
	\mathbf{w} = \text{Softmax}\left( \mathbf{W}_2 \cdot \text{ReLU}(\mathbf{W}_1 \cdot \text{GAP}(\mathbf{F}_{freq})) \right)
\end{equation}
Let $w_i$ denote the $i$-th channel of $\mathbf{w}$ ($i\in\{1,2,3\}$). A bank of depth-wise convolutions $\text{DWConv}_{k_i}(\cdot)$ with kernel sizes $k \in \{3, 5, 7\}$ extracts multi-scale responses from $\mathbf{F}_{freq}$. The adaptively weighted outputs are fused with the original input $\mathbf{F}_{in}$ via a skip connection, then refined by a point-wise convolution $\text{PWConv}(\cdot)$ with its own residual:
\begin{gather}
	\mathbf{F}_{sum} = \sum_{i=1}^{3} w_i \odot \text{DWConv}_{k_i}(\mathbf{F}_{freq}) + \mathbf{F}_{in} \\
	\mathbf{F}_{multi} = \text{PWConv}(\mathbf{F}_{sum}) + \mathbf{F}_{sum}
\end{gather}

Finally, $\mathbf{F}_{multi} \in \mathbb{R}^{B \times d \times H \times W}$ is spatially flattened back to a token sequence $\mathbf{Z}_{out} \in \mathbb{R}^{B \times N \times d}$, then projected to the original dimension $D$:
\begin{equation}
	\mathbf{H} = \mathbf{Z} + \left( \text{Dropout}(\text{GELU}(\mathbf{Z}_{out})) \mathbf{W}_{up} + \mathbf{b}_{up} \right)
\end{equation}
where $\mathbf{W}_{up} \in \mathbb{R}^{d \times D}$ is the up-projection matrix, and $\mathbf{Z}$ is carried as a residual so that the frozen block outputs are preserved. With $D{=}768$ and $d{=}64$, each HT module adds 113\,K parameters, totaling around 1.36\,M across all 12 layers. For reference, LoRA ($r{=}16$) adds about 1.18\,M parameters under the same backbone; unlike LoRA, which modifies attention weights inside each block, HT is appended after the block and introduces 2D spatial convolutions through FF and NE for image-domain processing.

\subsection{Downstream Tasks}

In this study, the downstream tasks comprise lesion segmentation and characteristics classification. To leverage the rich representations from the pre-trained encoder, we implement a multi-scale feature aggregation strategy followed by lightweight task-specific heads. 


\textbf{Multi-level Feature Aggregation}.
Rather than utilizing only the final embedding and inspired by CLIPSeg~\cite{clipseg_luddecke_2022}, we extract feature maps from intermediate Transformer blocks (specifically indices 3, 6, and 9) to capture a hierarchy of visual concepts. Let $\mathbf{F}_l$ denote the feature map from layer $l$. Each $\mathbf{F}_l$ is projected to a reduced dimension $d_{red}$ and refined through a non-linear block consisting of LayerNorm and linear layers. The processed features are then summed element-wise to yield a unified representation $\mathbf{F}_{agg} \in \mathbb{R}^{B \times d_{red} \times H \times W}$. This approach fuses low-level spatial details with high-level semantics while maintaining computational efficiency.

\textbf{Segmentation Head}.
For the segmentation task, we employ a lightweight decoding scheme to generate dense predictions. The unified feature map $\mathbf{F}_{agg}$ is bi-linearly up-sampled to the original image resolution ($224 \times 224$), followed by a $1 \times 1$ convolution that projects features into the label space. This minimalist design circumvents the computational burden of heavy transposed convolutions, ensuring parameter efficiency while delivering precise segmentation masks.

\textbf{Classification Head}.
Parallel to segmentation, the classification head extracts global descriptors from $\mathbf{F}_{agg}$. We apply Adaptive Average Pooling to collapse spatial dimensions, resulting in a compact feature vector. This vector is processed by a Multi-Layer Perceptron (MLP) consisting of two linear layers with ReLU activation and dropout. Such a structure balances representational capacity with regularization, minimizing the risk of over-fitting on data-scarce medical benchmarks.

%% file: sec/4_setup.tex
\section{Experimental Setup}

\subsection{Datasets}

To comprehensively evaluate the effectiveness of the proposed method, we utilize a diverse set of datasets for fine-tuning, segmentation, and classification tasks. An overview of the datasets and their partitions is provided in~\cref{tab:datasets}.

\input{tab/datasets}

\textbf{\textit{Fine-tuning Datasets.}} \textbf{PMC-OA}~\cite{pmcclip_lin_2023}, \textbf{CURD}~\cite{curd_li_2024}, and \textbf{MedPix}~\cite{medpix_siragusa_2025} serve as the primary data sources for the fine-tuning process. The original PMC-OA dataset contains 1,848,719 images across various medical imaging modalities. From this collection, ultrasound images and their corresponding captions were extracted, while low-resolution images were excluded. The CURD dataset comprises ultrasound images and captions from 7,390 patients presenting with breast, liver, or thyroid abnormalities. In addition, the MedPix dataset, which comprises 24,485 image-text pairs across CT and MRI imaging modalities, serves to bridge the gap between natural images and medical images, thereby facilitating the fine-tuning process for large models. The Qwen3-8B~\cite{qwen3_yang_2025} model was employed to translate captions in the CURD dataset into English, as they were originally written in Chinese.

During fine-tuning, the raw data underwent a multi-step preprocessing pipeline to ensure quality and consistency. First, we cleaned the textual captions by removing special characters and stripping extraneous whitespace. To provide meaningful descriptive textual content, samples with captions shorter than 20 characters were discarded. Following text cleaning, we removed any entries for which the corresponding image file was not found. The filtered subsets from these three sources were directly concatenated to form a unified fine-tuning corpus, from which samples were drawn uniformly at random during training. Finally, the resulting cleaned dataset was then randomly shuffled and partitioned into training and validation sets, using a 9:1 split.

\textbf{\textit{Downstream Task Datasets.}} We collected two \textbf{Lymph Node (LN)} datasets from two medical centers (Centers 1 and 2): \textbf{LN-INT} (Center 1) comprises 1,273 ultrasound images (785 benign and 488 malignant) of 307 cases, with years ranging from 2016 to 2024; and \textbf{LN-EXT} (Center 2) includes 374 images (190 benign and 184 malignant). In Center 1, 528 LNs were from newly diagnosed patients without prior treatment, and 745 were from follow-up cases with a history of radiotherapy or chemotherapy. Regarding anatomical distribution, LNs from Center 1 were predominantly located in the neck (1,253/1,273, 98.4\%), with a small number from the axillary ($n=18$) and inguinal ($n=2$) regions. In contrast, Center 2 exhibited greater anatomical diversity, including 284 neck, 71 axillary, 15 inguinal, and 4 popliteal LNs. In Center 2, 328 LNs were newly diagnosed, and 46 were follow-up cases. All images are annotated at the pixel level by experienced radiologists to delineate the LN region. Notably, \textbf{LN-EXT} is used exclusively as an external test set and is not involved in any part of the training process.

\textbf{BUSI}~\cite{busi_al_2020}. There are 780 breast lesion ultrasound images in BUSI, with 437 benign, 210 malignant, and 133 normal cases. The normal cases without ROIs are excluded in the following experiments as the purpose of this study is to detect and segment the abnormal regions.

\textbf{DDTI}~\cite{ddti_pedraza_2015}. The DDTI dataset consists of 637 ultrasound images of thyroid nodules. The dataset includes a wide range of lesion types including focal thyroiditis, cystic nodules, adenomas, and thyroid cancer.

\textbf{TN3K}~\cite{tn3k_gong_2023}. The TN3K dataset was collected at Zhujiang Hospital, Southern Medical University. For each patient, only one representative image is retained from those captured at similar viewpoints or anatomical locations. The dataset comprises 2,879 images for training and 614 images for testing.

\textbf{Prostate}~\cite{microsegnet_jiang_2023}. This dataset comprises both expert and non-expert prostate annotations of 75 patients, and only the expert annotations are used in this study.

\subsection{Implementation Details}

Prior to training, to minimize the impact of irrelevant information, non-imaging regions containing overlay annotations are cropped from self-collected ultrasound images. Subsequently, all images are resized to 224$\times$224 pixels for both training and testing. Except for TN3K~\cite{tn3k_gong_2023} and Prostate~\cite{microsegnet_jiang_2023}, all datasets are randomly divided into training, validation, and test sets in an 8:1:1 ratio. For TN3K and Prostate, the predefined training set is further split into training and validation subsets using an 8:2 ratio, while the original test set remains unchanged.

The whole fine-tuning process lasts for 32 epochs, while the models are trained for 200 epochs for the downstream tasks. The default learning rate for both fine-tuning and downstream tasks was set to $10^{-4}$, using a cosine annealing scheduler and a weight decay of $10^{-2}$. The loss functions used are InfoNCE loss for fine-tuning, Dice cross-entropy loss for segmentation, and focal loss for classification. All experiments utilize the AdamW optimizer with $\beta_1=0.9$ and $\beta_2=0.95$, and the batch size is set to 32. All experiments were conducted on a single NVIDIA 4090D GPU with 24 GB memory.

\subsection{Baselines and Comparison Methods}

The SOTA CLIP variants used for comparison in this study include raw CLIP~\cite{clip_radford_2021}, BiomedCLIP~\cite{biomedclip_zhang_2023}, UniMedCLIP~\cite{unimedclip_khattak_2024}, and MetaCLIP~\cite{metaclip_xu_2023}. To establish a visual-only performance baseline, we also evaluate models that rely solely on image input, including UNet~\cite{unet_ronneberger_2015}, ResNet18~\cite{resnet_he_2016}, and DINOv2~\cite{dinov2_oquab_2023}. For all experiments and comparisons, we adopt the ViT-Base architecture with a patch size of 16$\times$16 pixels as the standard backbone. Pre-trained weights are used when publicly available for the respective models.

To ensure the validity of the comparisons, the proposed head, which incorporates feature map up-sampling and adaptive average pooling for downstream tasks, was employed for all CLIP variants. Furthermore, all competitive methods utilize their own segmentation or classification heads to ensure a fair comparison. It is imperative to acknowledge that LN-EXT is exclusively utilized for evaluation purposes; consequently, the test results for LN-EXT are derived directly from the model that was trained using LN-INT.

\subsection{Prompt for Zero-shot Classification}

Given that the zero-shot classification inference process of CLIP~\cite{clip_radford_2021} necessitates text information as guidance, we devise specialized prompts for the ultrasound image classification task. Instead of adopting generic templates (\eg, ``a photo of a [CLASS]"), we construct an ensemble of clinically detailed descriptions for benign and malignant lesions, with each category comprising ten diverse prompts. These prompts capture specific sonographic features such as shape (oval vs. round), margin (circumscribed vs. spiculated), and echogenicity (homogeneous vs. heterogeneous). For instance, for lymph node classification, examples include ``A benign lymph node with an oval shape and a preserved fatty hilum'' and ``A malignant lymph node with loss of the central fatty hilum''. For breast lesions, we use prompts like ``A benign nodule with a parallel orientation, appearing wider-than-tall'' and ``A malignant nodule causing posterior acoustic shadowing''. The complete list of ensembled prompts is provided in~\cref{sec:appendix_prompts}. During inference, we compute the cosine similarity between the image embedding and the ensemble of text embeddings for each class. The final prediction is determined by averaging the similarity scores across all prompts within each class, thereby robustly aggregating multiple diagnostic criteria.

\subsection{Evaluation Metrics}

We used four widely-used metrics to evaluate the segmentation performance of the proposed method: Dice score (Dice, \%), intersection over union (IoU, \%), 95-percentile of Hausdorff distance (HD95), and average symmetric surface distance (ASD). For classification, Accuracy (Acc, \%), Recall (Rec, \%), Precision (Pre, \%), F1 score (\%), and area under receiver operating characteristic curve (AUC, \%) are adopted as evaluation metrics. All metrics are calculated based on the binary segmentation and classification results and averaged across all images in the test set.

%% file: tab/datasets.tex
\begin{table}[htb]
	\caption{Overview of datasets and partitions. Ft., Seg., and Cls. refer to fine-tuning, segmentation and classification tasks, respectively.}
	\centering
	\scriptsize
	\setlength{\tabcolsep}{3.5pt}
	\begin{tabular}{l|rrrr|ccc}
		\toprule
		Dataset                                & Training & Validation & Testing & Total  & Ft.        & Seg.       & Cls.       \\
		\midrule
		PMC-OA~\cite{pmcclip_lin_2023}         & 4,942    & 534        & 0       & 5,476  & \checkmark &            &            \\
		CURD~\cite{curd_li_2024}               & 13,228   & 1,500      & 0       & 14,728 & \checkmark &            &            \\
		MedPix~\cite{medpix_siragusa_2025}     & 21,512   & 2,391      & 0       & 23,910 & \checkmark &            &            \\
		\midrule
		LN-INT                                 & 1,019    & 127        & 127     & 1,273  &            & \checkmark & \checkmark \\
		LN-EXT                                 & 0        & 0          & 374     & 374    &            & \checkmark & \checkmark \\
		BUSI~\cite{busi_al_2020}               & 518      & 65         & 64      & 647    &            & \checkmark & \checkmark \\
		DDTI~\cite{ddti_pedraza_2015}          & 510      & 64         & 63      & 637    &            & \checkmark &            \\
		TN3K~\cite{tn3k_gong_2023}             & 2,304    & 575        & 614     & 3,493  &            & \checkmark &            \\
		Prostate~\cite{microsegnet_jiang_2023} & 1,937    & 215        & 758     & 2,910  &            & \checkmark &            \\
		\bottomrule
	\end{tabular}
	\label{tab:datasets}
\end{table}

%% file: sec/5_results.tex
\section{Results and Analysis}

To validate the effectiveness of the proposed method, we present a comprehensive evaluation covering segmentation, classification, few-shot learning, and cross-dataset generalization. We also provide ablation studies and efficiency analysis to further dissect the contribution of each component.

\subsection{Segmentation Performance}

The quantitative segmentation results are presented in~\cref{tab:results_seg}. It is evident that the proposed HT strategy consistently enhances performance across all datasets and CLIP variants. Notably, applying HT to the vanilla CLIP~\cite{clip_radford_2021} yields substantial improvements, achieving the highest Dice scores on LN-EXT (78.62\%), BUSI (81.19\%), TN3K (81.42\%), and Prostate (92.11\%) datasets. Similarly, MetaCLIP (HT) demonstrates top-tier performance on LN-INT (80.66\%) and DDTI (90.42\%).

\input{tab/results_seg}

When comparing HT-adapted models to their unadapted baselines, the performance gains are significant. For instance, on the challenging LN-INT dataset, CLIP (HT) surpasses the vanilla CLIP~\cite{clip_radford_2021} by 9.62\% in Dice score (80.55\% vs. 70.93\%), and MetaCLIP (HT) improves upon MetaCLIP~\cite{metaclip_xu_2023} by 10.48\% (80.66\% vs. 70.18\%). This trend of improvement is consistent across all metrics (IoU, HD95, ASD), with HT models exhibiting sharper boundary delineation as evidenced by lower HD95 and ASD scores. These findings indicate that HT can learn robust, transferable ultrasound representations even on small datasets.

Three CLIP variants were evaluated: BiomedCLIP~\cite{biomedclip_zhang_2023}, UniMedCLIP~\cite{unimedclip_khattak_2024}, and MetaCLIP~\cite{metaclip_xu_2023}. While trained on large-scale internet and medical data, none of these domain-specific models in their original form consistently surpass or match the performance of HT-enhanced models and CLIPSeg~\cite{clipseg_luddecke_2022}. We attribute this to the modality gap: although trained on general and confounding medical image-text data, their representations may not generalize directly to the heterogeneous acoustic signatures of ultrasound without specific adaptation. However, when equipped with the HT adapter, these models show significant performance boosts, validating the effectiveness of the proposed method.

\begin{figure*}[pos=htb]
	\centering
	\includegraphics[width=\linewidth]{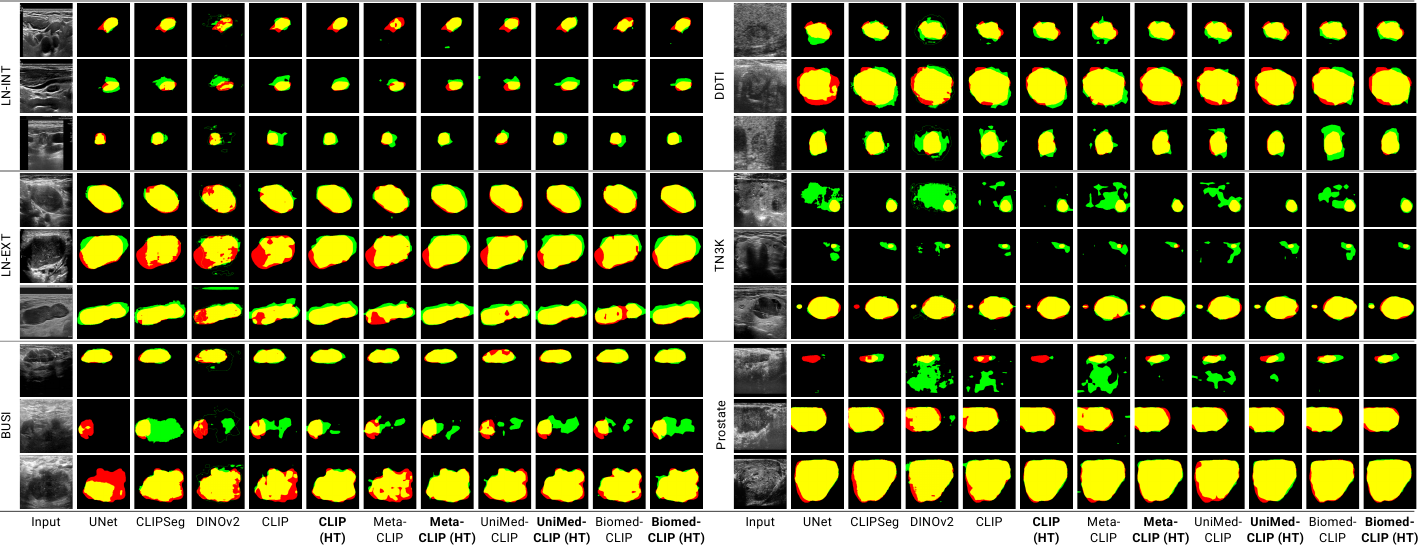}
	\caption{Segmentation visualization on LN-INT, LN-EXT, BUSI~\cite{busi_al_2020}, DDTI~\cite{ddti_pedraza_2015}, TN3K~\cite{tn3k_gong_2023}, and Prostate~\cite{microsegnet_jiang_2023} datasets. The first column shows the input images. Regions in red, green, and yellow indicate \textcolor{red}{false negative}, \textcolor{green}{false positive}, and \textcolor{YellowOrange}{true positive} predictions, respectively. Hybrid-tuned models are in bold.}
	\label{fig:seg-viz}
\end{figure*}

We randomly selected three samples per test set of the six datasets for visualization, as shown in~\cref{fig:seg-viz}. A thorough examination reveals that the proposed HT adaptation technique exhibits superior performance compared to other methods across all datasets, especially on the TN3K~\cite{tn3k_gong_2023} and Prostate~\cite{microsegnet_jiang_2023} datasets where it accurately delineates complex boundaries of thyroid nodule and prostate compared to baselines. These visual results align with the quantitative analysis.

Models equipped with HT demonstrated a marked improvement over the baselines, as evidenced by the second and third samples in the external test set (LN-EXT). The employment of HT models in depicting anatomical regions is characterized by a tendency to incorporate comprehensive and holistic approaches. This tendency is manifested through the strategic filling of internal voids and the refinement of lesion boundaries, thereby facilitating a more comprehensive depiction of the anatomical structure. Furthermore, although HT increases false positives (green area) while optimizing the true positives (yellow area) in the second sample from the BUSI~\cite{busi_al_2020} dataset, a divergent behavior is observed in other samples. This phenomenon is most evident in the last sample from the DDTI~\cite{ddti_pedraza_2015}, the first sample from the TN3K~\cite{tn3k_gong_2023}, and the Prostate~\cite{microsegnet_jiang_2023} datasets. The utilization of HT has been demonstrated to be a highly effective method for eliminating false positives in a substantial proportion of samples.

Finally, as demonstrated in~\cref{tab:results_seg} and~\cref{fig:seg-viz}, DINOv2~\cite{dinov2_oquab_2023} exhibited the poorest performance, even compared to the lightweight UNet. Despite being trained on a massive dataset (142 million images), it may lack the dense semantic support from text during pre-training. Consequently, it struggles with the domain gap between natural and medical ultrasound imaging data, leading to high false positive rates (\eg, in TN3K~\cite{tn3k_gong_2023} and Prostate~\cite{microsegnet_jiang_2023} datasets) and notable uncertainties in prediction.

\subsection{Classification Performance}

In this study, we evaluated a range of CLIP-based models and the HT strategy on three ultrasound classification datasets (LN-INT, LN-EXT, and BUSI~\cite{busi_al_2020}), as only they were labeled with diagnostic classes (\ie, benign and malignant) according to the confirmed pathology results. The quantitative results for the zero-shot and supervised classification are illustrated in~\cref{tab:results_cls}.

\input{tab/results_cls}

In the zero-shot setting, most CLIP variants perform near random guessing, indicating a lack of domain alignment. Adding HT generally decreases zero-shot accuracy, which can be attributed to a prediction bias shift rather than feature degradation. The contrastive fine-tuning on pathology-rich data shifts embeddings towards the malignant class, lowering accuracy but often maintaining or improving the rank-based AUC (\eg, CLIP on LN-INT: +5.39\%; LN-EXT: +6.73\%), effectively preserving discriminative capability. Since HT targets supervised adaptation, this bias is naturally corrected by the downstream classifier.

For the supervised classification task, the proposed HT strategy demonstrates a consistent performance improvement across various CLIP-based models. As presented in~\cref{tab:results_cls}, UniMedCLIP (HT) achieves the highest performance among CLIP variants, with an average accuracy of 83.04\% and an AUC of 91.44\%, surpassing the vanilla UniMedCLIP~\cite{unimedclip_khattak_2024}. Notably, BiomedCLIP (HT) also shows significant gains, improving the average accuracy and AUC by 3.75\% and 3.55\%, respectively, compared to its baseline. These results highlight the capability of HT to effectively adapt general medical vision-language models to the ultrasound domain.

In comparison with traditional deep learning models, ResNet18~\cite{resnet_he_2016} achieves a competitive average accuracy of 84.24\%. However, it is important to note that ResNet18 involves a large number of parameters that need to be trained. In contrast, the HT strategy allows CLIP variants to achieve comparable performance (\eg, UniMedCLIP (HT) vs. ResNet18) with significantly fewer trainable parameters (\ie, 4.1 M vs. 11.2 M), demonstrating superior parameter efficiency. On the other hand, DINOv2~\cite{dinov2_oquab_2023} exhibits suboptimal performance with an average accuracy of 57.57\%, which is considerably lower than both ResNet18 and the HT-enhanced CLIP models. This disparity further emphasizes the advantage of the proposed method in efficiently leveraging pre-trained representations for downstream tasks.

\subsection{Few-shot Learning}

\begin{figure}[pos=htb]
	\centering
	\includegraphics[width=\linewidth]{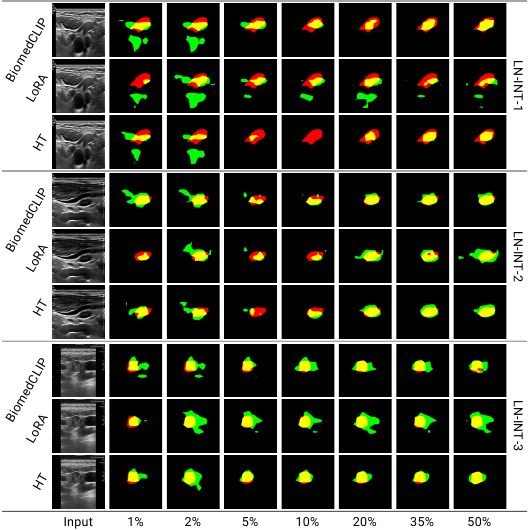}
	\caption{Few-shot segmentation visualization on LN-INT dataset. The first column shows the input images. The percentages represent the amount of data used for model training. Regions in red, green, and yellow indicate \textcolor{red}{false negative}, \textcolor{green}{false positive}, and \textcolor{YellowOrange}{true positive} predictions, respectively.}
	\label{fig:few-viz}
\end{figure}

To investigate the data efficiency of the proposed method, a series of few-shot learning experiments were conducted on both segmentation and classification tasks across all datasets. Ratio-based sampling is adopted rather than a fixed k-shot setting because the datasets differ substantially in size and the training data ratio varies from 1\% to 50\%. The performance of BiomedCLIP~\cite{biomedclip_zhang_2023} and its adaptation variants (LoRA~\cite{lora_hu_2022} and the proposed HT) are compared. The quantitative outcomes and the segmentation visualizations are presented in~\cref{tab:results_few_shot} and~\cref{fig:few-viz}, respectively.

\input{tab/results_few_shot}

In terms of segmentation, the HT strategy demonstrates superior data efficiency and performance stability. Even with extremely limited data (1\% ratio), HT achieves competitive Dice scores (63.55\%) compared to the baseline BiomedCLIP~\cite{biomedclip_zhang_2023} (63.98\%) while significantly outperforming it in boundary metrics (HD95: 39.05 vs. 41.18; ASD: 14.64 vs. 15.58). As the training data ratio increases, the advantage of HT becomes more pronounced. From 2\% to 50\% ratios, HT consistently achieves the best performance across all segmentation metrics. For instance, at a 20\% ratio, HT surpasses BiomedCLIP by 3.29\% in Dice and reduces HD95 by 8.37, indicating that the HT module effectively captures detailed anatomical structures and boundaries with limited supervision. Furthermore, HT consistently outperforms LoRA~\cite{lora_hu_2022}, suggesting that the hybrid tuning strategy is more effective than standard PEFT for dense prediction tasks in ultrasound.

For classification, the HT model also exhibits robust performance. At the 1\% ratio, HT achieves the highest AUC (65.27\%) and F1 score (32.93\%), demonstrating its capability to learn discriminative features from scarce data. While the performance fluctuates slightly at lower ratios (2\%-5\%), HT establishes a clear lead as the data volume increases. At 20\%, 35\%, and 50\% ratios, it consistently secures the highest Accuracy, Recall, F1, and AUC scores. Notably, at the 50\% ratio, HT achieves an accuracy of 77.16\% and an AUC of 86.51\%, significantly outperforming both the baseline BiomedCLIP~\cite{biomedclip_zhang_2023} and LoRA~\cite{lora_hu_2022}. This trend confirms that the HT strategy not only adapts well to the domain but also scales effectively with available data, mitigating the risk of over-fitting or catastrophic forgetting often observed in zero-shot and few-shot scenarios.

\subsection{Cross-dataset Generalization}

\begin{figure*}[pos=htb]
	\centering
	\includegraphics[width=0.95\linewidth]{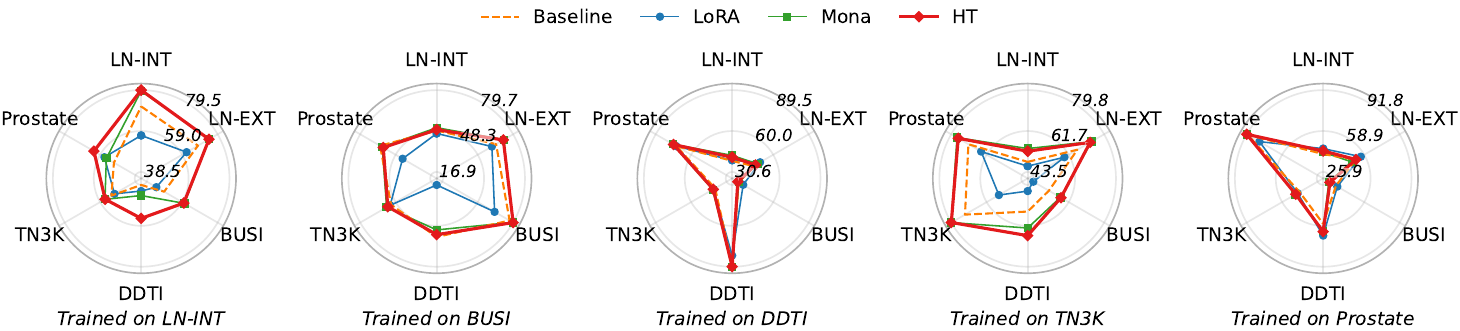}
	\caption{Overview of Dice score for cross-dataset segmentation evaluation. The numbers from inside to outside of radar show the minimum, average and maximum values of each experiment.}
	\label{fig:seg-radar}
\end{figure*}

To further demonstrate the robustness and generalization capability of the proposed HT, comprehensive cross-dataset evaluations were conducted. We adopted a leave-one-domain-out strategy, where models were trained on one dataset and evaluated on all other datasets. This rigorous setting challenges the models to handle significant domain shifts arising from different anatomical structures (\eg, thyroid, breast, lymph node, prostate) and acquisition protocols. Evaluation results are summarized in~\cref{tab:cross_evaluation} and visualized in~\cref{fig:seg-radar}.

\input{tab/cross_evaluation}

As detailed in~\cref{tab:cross_evaluation} and~\cref{fig:seg-radar}, the HT strategy demonstrates superior generalization in segmentation tasks. It achieves the highest Dice scores across all evaluation scopes: 84.03\% for in-domain, 55.50\% for cross-domain, and 60.25\% overall. This consistent outperformance against the baseline (Overall Dice: 56.14\%) and the SOTA Mona~\cite{mona_yin_2024} adapter (Overall Dice: 59.52\%) validates the effectiveness of HT in learning robust, shape-aware representations. Conversely, LoRA~\cite{lora_hu_2022} exhibits negative transfer, with its overall Dice score dropping to 51.74\%, significantly lower than the baseline, confirming that standard PEFT methods may struggle with dense prediction tasks under domain shifts.

\begin{figure}[pos=htb]
	\centering
	\includegraphics[width=0.725\linewidth]{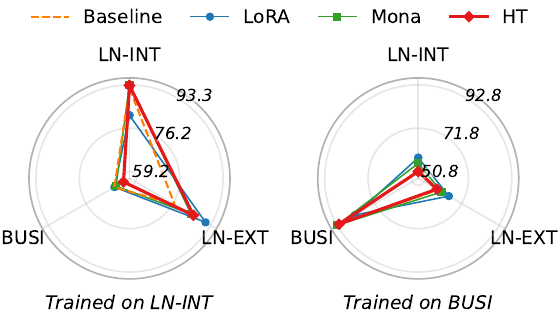}
	\caption{Overview of AUC for cross-dataset classification evaluation. The numbers from inside to outside of radar show the minimum, average and maximum values of each experiment.}
	\label{fig:cls-radar}
\end{figure}

For classification, as illustrated in~\cref{tab:cross_evaluation} and~\cref{fig:cls-radar}, HT achieves the leading overall Accuracy of 69.42\% and an AUC of 73.05\%, surpassing the baseline (Acc: 66.89\%, AUC: 72.12\%). Notably, while LoRA demonstrates strong cross-domain transferability in classification metrics (AUC: 69.10\%), its in-domain performance (Acc: 67.82\%) lags significantly behind HT (Acc: 86.28\%). This suggests a trade-off between specificity and generalization in standard fine-tuning methods. In contrast, HT ensures high diagnostic precision on internal data while improving generalization to heterogeneous external datasets, leveraging the HT module to mitigate the risk of over-fitting and maintain a balanced performance profile.

\subsection{Ablation Studies}

To verify the effectiveness of the proposed method, we conducted comprehensive ablation studies on segmentation, classification, and the internal components of the HT module based on BiomedCLIP~\cite{biomedclip_zhang_2023}.

\textbf{Effectiveness on Segmentation.} We first evaluate the impact of different tuning strategies on segmentation performance, as reported in~\cref{tab:abla_seg}. The proposed HT method achieves the highest average Dice score of 82.44\% and IoU of 73.53\%, consistently outperforming other approaches. A notable observation is the performance of LoRA-based methods. Despite being a popular PEFT technique, LoRA~\cite{lora_hu_2022} exhibits a significant performance degradation compared to the baseline (\eg, Average Dice: 66.37\% vs. 78.02\%). Even when fine-tuning both visual and text encoders (LoRA V+T) to align the modalities, the performance remains suboptimal (66.64\%). This suggests that modifying the internal weights of the pre-trained encoders on limited ultrasound data may disrupt the learned general medical representations, leading to negative transfer. In contrast, adapter-based methods like Mona~\cite{mona_yin_2024} and HT, which freeze the backbone and learn additional adaptation layers, effectively preserve the pre-trained knowledge while adapting to the specific domain, resulting in superior segmentation accuracy.

\input{tab/abla_seg}

\textbf{Effectiveness on Classification.} A similar trend is observed in the classification task, as shown in~\cref{tab:abla_cls}. The HT strategy secures the top performance with an average accuracy of 80.81\% and an AUC of 90.14\%. While the Mona~\cite{mona_yin_2024} adapter also improves over the baseline (78.84\% vs. 77.06\%), the LoRA~\cite{lora_hu_2022} variants again fail to yield positive gains. Specifically, LoRA and LoRA (V+T) result in average accuracies of 68.92\% and 70.17\%, respectively, both falling below the baseline. This consistent underperformance of LoRA across tasks further confirms that preserving the integrity of the pre-trained backbone is crucial for small-scale ultrasound datasets. The HT strategy successfully bridges the domain gap without compromising the robust feature extraction capabilities of the original model.

\input{tab/abla_cls}

\textbf{Component Analysis of HT Module.} Finally, we analyze the contribution of individual components within the HT module: the Noise Estimation (NE) and Frequency Filtering (FF) blocks. As presented in~\cref{tab:abla_ht_module}, we compare the full HT model with variants using only NE or FF, as well as the Mona~\cite{mona_yin_2024} adapter. The results demonstrate that both components contribute to the overall performance. The Full HT model achieves the best results in both segmentation and classification. Notably, the NE module appears to play a pivotal role; using NE alone yields an average classification accuracy of 80.38\%, which is higher than using FF alone (79.51\%) and Mona (78.84\%). This indicates that estimating and mitigating the noise or uncertainty in the pre-trained features is essential for robust ultrasound image analysis. Regarding boundary accuracy, enabling NE changes HD95 by only 0.09 (19.33 to 19.42) and ASD by 0.01 (6.70 to 6.71) compared to the FF-only variant, confirming that the adaptive multi-scale smoothing of NE does not degrade boundary delineation. The combination of NE and FF (Full) further enhances the performance, validating the synergistic effect of the proposed hybrid tuning architecture.

\input{tab/abla_ht_module}

\begin{figure}[pos=htb]
	\centering
	\includegraphics[width=\linewidth]{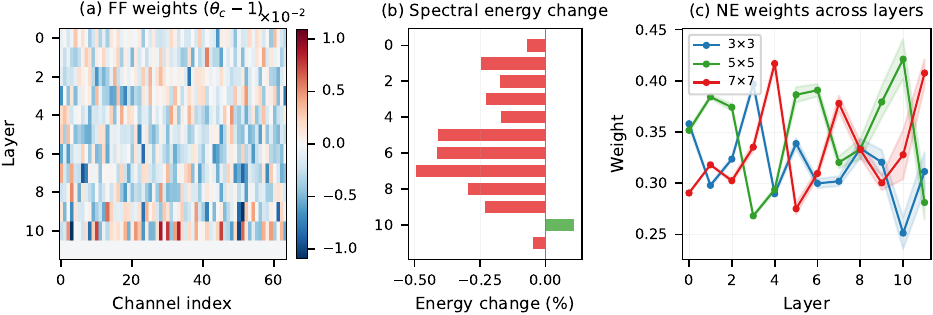}
	\caption{Spectral analysis of FF/NE modules across 12 layers. (a) Deviation of learned FF weights from initialization ($\theta_c-1$). (b) Layer-wise spectral energy change after FF. (c) NE attention weights showing layer-dependent multi-scale adaptation.}
	\label{fig:spectral-analysis}
\end{figure}

\textbf{Spectral Analysis of FF and NE Modules.} To provide mechanistic evidence, we probe the bottleneck feature maps of the trained HT model across all 12 transformer layers on test images drawn equally from BUSI, LN-INT, and LN-EXT (\cref{fig:spectral-analysis}). For the FF module, the learned channel-wise frequency filter weights $\theta_c$ (initialized at 1.0) exhibit a subtle but statistically significant downward shift after training ($0.9989\pm0.0027$, $p<0.001$), producing consistent spectral energy reduction across 11 of 12 layers (avg.\ $-0.22\%$; \cref{fig:spectral-analysis}a,b). Since each bottleneck channel encodes distinct frequency patterns, this channel-wise gating effectively performs soft frequency-band selection. For the NE module, the adaptive weights display a clear layer-dependent pattern (\cref{fig:spectral-analysis}c): in early layers, the fine-scale $3\times3$ kernel dominates ($w_{3\times3}=0.358$), while in late layers, the coarse $7\times7$ kernel prevails ($w_{7\times7}=0.408$); the medium $5\times5$ kernel is most frequently dominant overall (6/12 layers), matching the characteristic spatial scale of ultrasound speckle. These findings confirm that FF provides frequency-domain regularization and NE adaptively balances detail preservation versus noise suppression according to the feature abstraction level.

\textbf{Statistical Significance.} To rigorously validate the observed improvements, we conduct paired $t$-tests across all comparisons. For the main segmentation results (\cref{tab:results_seg}), HT vs.\ Baseline comparisons are statistically significant ($p<0.05$) in 54 out of 56 combinations, with 46 reaching $p<0.01$. Per-backbone average Dice improvements are all highly significant: CLIP ($\Delta\text{Dice}=+5.58$, $p<0.001$), MetaCLIP ($\Delta\text{Dice}=+6.29$, $p<0.001$), UniMedCLIP ($\Delta\text{Dice}=+4.33$, $p=0.001$), and BiomedCLIP ($\Delta\text{Dice}=+4.42$, $p=0.002$). For cross-dataset generalization, HT significantly outperforms Baseline (Overall Dice $p=0.007$) and LoRA ($p=0.008$). The HT vs.\ Mona~\cite{mona_yin_2024} comparison yields consistent but modest improvements ($\Delta\text{Dice}=+0.73$, $p=0.34$). The advantage of HT over Mona becomes more evident under cross-domain shifts, where HT achieves a $+0.84\%$ Dice improvement with lower variance across all folds.

\subsection{Inference Efficiency}

To assess clinical practicality, we compare inference efficiency across all adapter methods (\cref{tab:efficiency}). HT and Mona~\cite{mona_yin_2024} share identical GFLOPs (14.99G) since the FF and NE modules introduce negligible computational overhead (only $+$0.01\,M total parameters over Mona). While HT has a lower FPS (93.2) than the frozen Baseline (291.7) and Mona (120.1) due to the additional forward pass through the FF and NE branches, this corresponds to a per-image latency of 10.73\,ms, which is well within the real-time requirement for clinical ultrasound analysis ($>$30 FPS at batch inference). Notably, LoRA introduces substantially higher GFLOPs (19.79G, $+34\%$ over Baseline) while achieving inferior task performance, making HT a more favorable trade-off between efficiency and accuracy.

\input{tab/efficiency}

\subsection{Failure Case Analysis}
\label{sec:failure_cases}

\begin{figure}[pos=htb]
	\centering
	\includegraphics[width=\linewidth]{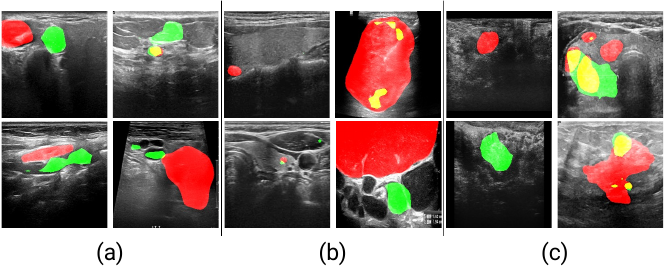}
	\caption{Visualization of typical failure cases: (a) insufficient discrimination between similar structures, (b) extreme scale variance in lesions, and (c) complex background echotexture. Regions in red, green, and yellow indicate \textcolor{red}{false negative}, \textcolor{green}{false positive}, and \textcolor{YellowOrange}{true positive} predictions, respectively.}
	\label{fig:failure_cases}
\end{figure}

While the proposed HT module demonstrates robust performance across multiple datasets, understanding its limitations is vital for clinical safety and practical deployment. We present visual examples of typical failure modes encountered by the proposed HT module in~\cref{fig:failure_cases}.

Specifically, we identify three primary failure modes. First, when adjacent anatomical structures share similar echogenic properties or morphological features with the target lesion, the model may struggle to accurately delineate boundaries, leading to merged segmentations or misclassification (\cref{fig:failure_cases}a). Second, extreme scale variance presents a significant challenge, where disproportionately small or large lesions can cause substantial under-segmentation or over-prediction errors (\cref{fig:failure_cases}b). Finally, severe heterogeneous background tissue can confuse the model, making it difficult to differentiate isoechoic lesions from surrounding fibroglandular tissue and resulting in false positive predictions (\cref{fig:failure_cases}c).

\subsection{Limitations and Future Perspectives}

While the proposed HT strategy demonstrates significant effectiveness, three limitations warrant further investigation.

First, foundational models like BiomedCLIP~\cite{biomedclip_zhang_2023} are pre-trained on broad biomedical datasets where ultrasound images and reports are underrepresented and often of lower quality compared to CT or MRI. Consequently, these models inherently lack robust encoding of ultrasound-specific artifacts (\eg, speckle noise, acoustic shadowing) and domain-specific terminologies. Developing a large-scale, high-quality, vision-language ultrasound dataset, akin to MS-COCO~\cite{coco_lin_2014} or ROCOv2~\cite{rocov2_ruckert_2024}, remains a critical prerequisite for advancing ultrasound-specific FMs.

Second, our zero-shot analysis reveals that contrastive fine-tuning on pathology-rich datasets induces a prediction bias shift. Specifically, image embeddings are pulled closer to malignant text descriptions, leading to over-prediction of the malignant class and unreliable accuracy metrics, despite preserving overall discriminative ability (AUC). Future research should address this calibration issue through weight-space ensembling~\cite{wiseft_wortsman_2022}, test-time calibration (\eg, temperature scaling), or retrieval-augmented classification to ground predictions in confirmed clinical cases.

Third, the suboptimal performance of joint fine-tuning (\eg, LoRA V+T) highlights a persistent misalignment between global CLIP representations and fine-grained ultrasound diagnostic criteria. Because HT primarily adapts visual features, it potentially under-utilizes rich clinical textual knowledge. Future efforts must explore learnable context optimization (\eg, prompt tuning) or robust textual adaptation mechanisms to achieve deeper, synergistic integration of vision and language modalities in ultrasound analysis.

%% file: tab/results_seg.tex
\begin{table*}[htb]
	\caption{Segmentation results on six ultrasound datasets, averaged over three runs with different random seeds. (HT) denotes hybrid-tuned models. Performance \textcolor{red}{improvements} over baselines are highlighted. Mean{\tiny$\pm$std} is reported.}
	\centering
	\scriptsize
	\setlength{\tabcolsep}{2pt}
	\begin{tabular}{lr|llll|llll|llll}
		\toprule
		\multirow{2}{*}{\textbf{Method}}  & \multirow{2}{*}{\textbf{Params}} & \multicolumn{4}{c|}{\textbf{LN-INT}}   & \multicolumn{4}{c|}{\textbf{LN-EXT}}   & \multicolumn{4}{c}{\textbf{BUSI}}                                                                                                                                                                                                                                                                                                                                                                                    \\ \cmidrule{3-14}
		                                  &                                  & \textbf{Dice}                          & \textbf{IoU}                           & \textbf{HD95}                          & \textbf{ASD}                          & \textbf{Dice}                          & \textbf{IoU}                           & \textbf{HD95}                          & \textbf{ASD}                          & \textbf{Dice}                          & \textbf{IoU}                           & \textbf{HD95}                          & \textbf{ASD}                          \\
		\midrule
		UNet                              & 1.8 M                            & 77.41{\tiny$\pm$0.54}                  & 68.36{\tiny$\pm$0.46}                  & 24.99{\tiny$\pm$1.45}                  & 9.03{\tiny$\pm$0.47}                  & 70.54{\tiny$\pm$1.17}                  & 59.82{\tiny$\pm$1.31}                  & 34.08{\tiny$\pm$2.60}                  & 11.25{\tiny$\pm$1.20}                 & 76.97{\tiny$\pm$1.74}                  & 67.11{\tiny$\pm$1.80}                  & 26.42{\tiny$\pm$2.60}                  & 8.45{\tiny$\pm$1.94}                  \\
		CLIPSeg                           & 1.1 M                            & 73.14{\tiny$\pm$0.20}                  & 62.31{\tiny$\pm$0.19}                  & 26.58{\tiny$\pm$0.72}                  & 10.16{\tiny$\pm$0.22}                 & 73.28{\tiny$\pm$0.15}                  & 62.61{\tiny$\pm$0.16}                  & 28.44{\tiny$\pm$0.32}                  & 9.88{\tiny$\pm$0.01}                  & 79.35{\tiny$\pm$0.44}                  & 69.24{\tiny$\pm$0.45}                  & 19.47{\tiny$\pm$0.65}                  & 6.65{\tiny$\pm$0.37}                  \\
		DINOv2                            & 9.7 M                            & 62.09{\tiny$\pm$1.01}                  & 49.51{\tiny$\pm$1.29}                  & 140.41{\tiny$\pm$15.81}                & 52.31{\tiny$\pm$5.19}                 & 56.54{\tiny$\pm$0.70}                  & 43.26{\tiny$\pm$0.66}                  & 147.87{\tiny$\pm$8.15}                 & 59.19{\tiny$\pm$3.08}                 & 61.29{\tiny$\pm$1.39}                  & 48.28{\tiny$\pm$1.61}                  & 177.75{\tiny$\pm$30.75}                & 72.43{\tiny$\pm$11.94}                \\
		\cmidrule{1-14}
		CLIP                              & 3.0 M                            & 70.93{\tiny$\pm$0.23}                  & 58.79{\tiny$\pm$0.37}                  & 35.01{\tiny$\pm$0.98}                  & 12.20{\tiny$\pm$0.58}                 & 71.34{\tiny$\pm$0.58}                  & 58.42{\tiny$\pm$0.72}                  & 37.14{\tiny$\pm$0.70}                  & 12.66{\tiny$\pm$0.11}                 & 77.31{\tiny$\pm$0.58}                  & 65.47{\tiny$\pm$0.70}                  & 29.84{\tiny$\pm$2.64}                  & 9.21{\tiny$\pm$1.04}                  \\
		\rowcolor{gray!10}CLIP (HT)       & 4.4 M                            & \textcolor{red}{80.55{\tiny$\pm$0.58}} & \textcolor{red}{71.94{\tiny$\pm$0.71}} & \textcolor{red}{18.44{\tiny$\pm$0.63}} & \textcolor{red}{6.81{\tiny$\pm$0.28}} & \textcolor{red}{78.62{\tiny$\pm$0.20}} & \textcolor{red}{68.20{\tiny$\pm$0.05}} & \textcolor{red}{22.69{\tiny$\pm$0.22}} & \textcolor{red}{7.60{\tiny$\pm$0.19}} & \textcolor{red}{81.19{\tiny$\pm$1.55}} & \textcolor{red}{71.25{\tiny$\pm$1.42}} & \textcolor{red}{17.84{\tiny$\pm$1.37}} & \textcolor{red}{5.45{\tiny$\pm$0.56}} \\
		MetaCLIP                          & 2.8 M                            & 70.18{\tiny$\pm$0.51}                  & 58.01{\tiny$\pm$0.82}                  & 40.24{\tiny$\pm$5.53}                  & 13.86{\tiny$\pm$1.69}                 & 68.82{\tiny$\pm$0.52}                  & 55.73{\tiny$\pm$0.48}                  & 41.58{\tiny$\pm$1.56}                  & 14.11{\tiny$\pm$0.63}                 & 76.12{\tiny$\pm$0.33}                  & 64.22{\tiny$\pm$0.36}                  & 30.47{\tiny$\pm$3.66}                  & 9.07{\tiny$\pm$0.78}                  \\
		\rowcolor{gray!10}MetaCLIP (HT)   & 4.1 M                            & \textcolor{red}{80.66{\tiny$\pm$0.20}} & \textcolor{red}{71.75{\tiny$\pm$0.05}} & \textcolor{red}{18.71{\tiny$\pm$1.61}} & \textcolor{red}{6.86{\tiny$\pm$0.39}} & \textcolor{red}{77.14{\tiny$\pm$0.24}} & \textcolor{red}{66.53{\tiny$\pm$0.29}} & \textcolor{red}{23.65{\tiny$\pm$1.17}} & \textcolor{red}{8.01{\tiny$\pm$0.31}} & \textcolor{red}{80.74{\tiny$\pm$0.33}} & \textcolor{red}{70.64{\tiny$\pm$0.36}} & \textcolor{red}{18.21{\tiny$\pm$1.76}} & \textcolor{red}{5.86{\tiny$\pm$0.72}} \\
		UniMedCLIP                        & 2.8 M                            & 73.05{\tiny$\pm$0.14}                  & 62.35{\tiny$\pm$0.16}                  & 32.33{\tiny$\pm$1.26}                  & 11.17{\tiny$\pm$0.58}                 & 68.68{\tiny$\pm$0.29}                  & 56.53{\tiny$\pm$0.16}                  & 34.69{\tiny$\pm$0.54}                  & 11.95{\tiny$\pm$0.05}                 & 75.33{\tiny$\pm$0.28}                  & 63.52{\tiny$\pm$0.24}                  & 26.72{\tiny$\pm$0.57}                  & 8.73{\tiny$\pm$0.13}                  \\
		\rowcolor{gray!10}UniMedCLIP (HT) & 4.1 M                            & \textcolor{red}{78.85{\tiny$\pm$0.22}} & \textcolor{red}{70.17{\tiny$\pm$0.34}} & \textcolor{red}{18.89{\tiny$\pm$1.57}} & \textcolor{red}{6.63{\tiny$\pm$0.56}} & \textcolor{red}{75.22{\tiny$\pm$0.18}} & \textcolor{red}{64.39{\tiny$\pm$0.36}} & \textcolor{red}{25.53{\tiny$\pm$0.46}} & \textcolor{red}{8.63{\tiny$\pm$0.33}} & \textcolor{red}{78.85{\tiny$\pm$0.85}} & \textcolor{red}{68.37{\tiny$\pm$1.03}} & \textcolor{red}{20.08{\tiny$\pm$1.82}} & \textcolor{red}{6.44{\tiny$\pm$0.35}} \\
		BiomedCLIP                        & 2.8 M                            & 71.27{\tiny$\pm$0.32}                  & 60.25{\tiny$\pm$0.89}                  & 34.41{\tiny$\pm$1.73}                  & 11.49{\tiny$\pm$0.65}                 & 68.07{\tiny$\pm$0.19}                  & 55.95{\tiny$\pm$0.06}                  & 34.72{\tiny$\pm$0.92}                  & 11.63{\tiny$\pm$0.33}                 & 76.57{\tiny$\pm$0.64}                  & 64.98{\tiny$\pm$0.72}                  & 29.35{\tiny$\pm$2.29}                  & 9.85{\tiny$\pm$0.52}                  \\
		\rowcolor{gray!10}BiomedCLIP (HT) & 4.1 M                            & \textcolor{red}{79.45{\tiny$\pm$0.24}} & \textcolor{red}{70.31{\tiny$\pm$0.29}} & \textcolor{red}{20.81{\tiny$\pm$0.26}} & \textcolor{red}{7.40{\tiny$\pm$0.15}} & \textcolor{red}{74.46{\tiny$\pm$0.98}} & \textcolor{red}{63.84{\tiny$\pm$0.94}} & \textcolor{red}{25.16{\tiny$\pm$0.58}} & \textcolor{red}{8.46{\tiny$\pm$0.18}} & \textcolor{red}{79.73{\tiny$\pm$0.66}} & \textcolor{red}{69.29{\tiny$\pm$0.67}} & \textcolor{red}{21.78{\tiny$\pm$1.44}} & \textcolor{red}{7.05{\tiny$\pm$0.30}} \\
		\midrule
		\multirow{2}{*}{\textbf{Method}}  & \multirow{2}{*}{\textbf{Params}} & \multicolumn{4}{c|}{\textbf{DDTI}}     & \multicolumn{4}{c|}{\textbf{TN3K}}     & \multicolumn{4}{c}{\textbf{Prostate}}                                                                                                                                                                                                                                                                                                                                                                                \\ \cmidrule{3-14}
		                                  &                                  & \textbf{Dice}                          & \textbf{IoU}                           & \textbf{HD95}                          & \textbf{ASD}                          & \textbf{Dice}                          & \textbf{IoU}                           & \textbf{HD95}                          & \textbf{ASD}                          & \textbf{Dice}                          & \textbf{IoU}                           & \textbf{HD95}                          & \textbf{ASD}                          \\
		\midrule
		UNet                              & 1.8 M                            & 86.44{\tiny$\pm$2.97}                  & 76.93{\tiny$\pm$4.11}                  & 21.11{\tiny$\pm$2.77}                  & 7.63{\tiny$\pm$0.92}                  & 79.49{\tiny$\pm$0.57}                  & 70.29{\tiny$\pm$0.67}                  & 24.21{\tiny$\pm$1.69}                  & 7.97{\tiny$\pm$0.32}                  & 91.64{\tiny$\pm$0.05}                  & 86.93{\tiny$\pm$0.13}                  & 14.27{\tiny$\pm$0.34}                  & 4.28{\tiny$\pm$0.12}                  \\
		CLIPSeg                           & 1.1 M                            & 87.89{\tiny$\pm$0.09}                  & 78.82{\tiny$\pm$0.14}                  & 18.29{\tiny$\pm$0.25}                  & 7.15{\tiny$\pm$0.11}                  & 77.19{\tiny$\pm$0.28}                  & 67.06{\tiny$\pm$0.37}                  & 23.16{\tiny$\pm$1.03}                  & 7.84{\tiny$\pm$0.49}                  & 91.32{\tiny$\pm$0.11}                  & 86.05{\tiny$\pm$0.16}                  & 14.90{\tiny$\pm$0.56}                  & 4.73{\tiny$\pm$0.14}                  \\
		DINOv2                            & 9.7 M                            & 86.20{\tiny$\pm$0.17}                  & 76.21{\tiny$\pm$0.29}                  & 54.93{\tiny$\pm$3.12}                  & 21.80{\tiny$\pm$1.19}                 & 60.71{\tiny$\pm$0.41}                  & 47.62{\tiny$\pm$0.47}                  & 124.10{\tiny$\pm$3.26}                 & 47.74{\tiny$\pm$1.81}                 & 87.52{\tiny$\pm$0.04}                  & 80.05{\tiny$\pm$0.08}                  & 85.79{\tiny$\pm$1.67}                  & 28.81{\tiny$\pm$0.77}                 \\
		\cmidrule{1-14}
		CLIP                              & 3.0 M                            & 87.56{\tiny$\pm$0.15}                  & 78.32{\tiny$\pm$0.22}                  & 19.63{\tiny$\pm$0.48}                  & 7.66{\tiny$\pm$0.17}                  & 72.89{\tiny$\pm$0.46}                  & 61.22{\tiny$\pm$0.45}                  & 36.32{\tiny$\pm$2.58}                  & 12.09{\tiny$\pm$0.95}                 & 90.53{\tiny$\pm$0.11}                  & 84.56{\tiny$\pm$0.14}                  & 23.41{\tiny$\pm$1.03}                  & 7.18{\tiny$\pm$0.13}                  \\
		\rowcolor{gray!10}CLIP (HT)       & 4.4 M                            & \textcolor{red}{90.11{\tiny$\pm$0.19}} & \textcolor{red}{82.39{\tiny$\pm$0.35}} & \textcolor{red}{15.39{\tiny$\pm$0.44}} & \textcolor{red}{5.80{\tiny$\pm$0.11}} & \textcolor{red}{81.42{\tiny$\pm$0.11}} & \textcolor{red}{71.81{\tiny$\pm$0.37}} & \textcolor{red}{18.65{\tiny$\pm$0.55}} & \textcolor{red}{6.09{\tiny$\pm$0.10}} & \textcolor{red}{92.11{\tiny$\pm$0.18}} & \textcolor{red}{87.34{\tiny$\pm$0.22}} & \textcolor{red}{11.67{\tiny$\pm$0.21}} & \textcolor{red}{3.87{\tiny$\pm$0.13}} \\
		MetaCLIP                          & 2.8 M                            & 88.16{\tiny$\pm$0.37}                  & 79.22{\tiny$\pm$0.59}                  & 19.71{\tiny$\pm$1.09}                  & 7.40{\tiny$\pm$0.32}                  & 71.19{\tiny$\pm$0.36}                  & 59.40{\tiny$\pm$0.56}                  & 40.43{\tiny$\pm$0.45}                  & 13.71{\tiny$\pm$0.19}                 & 89.92{\tiny$\pm$0.13}                  & 83.53{\tiny$\pm$0.12}                  & 26.97{\tiny$\pm$0.72}                  & 8.20{\tiny$\pm$0.41}                  \\
		\rowcolor{gray!10}MetaCLIP (HT)   & 4.1 M                            & \textcolor{red}{90.42{\tiny$\pm$0.29}} & \textcolor{red}{82.87{\tiny$\pm$0.47}} & \textcolor{red}{14.35{\tiny$\pm$0.71}} & \textcolor{red}{5.59{\tiny$\pm$0.19}} & \textcolor{red}{81.09{\tiny$\pm$0.44}} & \textcolor{red}{71.40{\tiny$\pm$0.45}} & \textcolor{red}{18.64{\tiny$\pm$0.15}} & \textcolor{red}{6.11{\tiny$\pm$0.40}} & \textcolor{red}{92.05{\tiny$\pm$0.06}} & \textcolor{red}{87.15{\tiny$\pm$0.10}} & \textcolor{red}{12.23{\tiny$\pm$0.53}} & \textcolor{red}{4.03{\tiny$\pm$0.16}} \\
		UniMedCLIP                        & 2.8 M                            & 88.73{\tiny$\pm$0.31}                  & 80.11{\tiny$\pm$0.49}                  & 18.00{\tiny$\pm$0.51}                  & 6.72{\tiny$\pm$0.09}                  & 71.51{\tiny$\pm$0.72}                  & 59.80{\tiny$\pm$0.79}                  & 34.95{\tiny$\pm$0.78}                  & 12.10{\tiny$\pm$0.28}                 & 90.36{\tiny$\pm$0.18}                  & 84.43{\tiny$\pm$0.24}                  & 18.88{\tiny$\pm$0.44}                  & 6.01{\tiny$\pm$0.18}                  \\
		\rowcolor{gray!10}UniMedCLIP (HT) & 4.1 M                            & \textcolor{red}{89.91{\tiny$\pm$0.06}} & \textcolor{red}{82.06{\tiny$\pm$0.03}} & \textcolor{red}{15.65{\tiny$\pm$0.58}} & \textcolor{red}{5.86{\tiny$\pm$0.07}} & \textcolor{red}{79.30{\tiny$\pm$0.45}} & \textcolor{red}{69.22{\tiny$\pm$0.49}} & \textcolor{red}{21.13{\tiny$\pm$0.91}} & \textcolor{red}{6.89{\tiny$\pm$0.26}} & \textcolor{red}{91.54{\tiny$\pm$0.07}} & \textcolor{red}{86.32{\tiny$\pm$0.13}} & \textcolor{red}{12.96{\tiny$\pm$0.34}} & \textcolor{red}{4.40{\tiny$\pm$0.07}} \\
		BiomedCLIP                        & 2.8 M                            & 88.55{\tiny$\pm$0.12}                  & 79.85{\tiny$\pm$0.18}                  & 18.49{\tiny$\pm$0.47}                  & 6.90{\tiny$\pm$0.10}                  & 72.76{\tiny$\pm$0.36}                  & 61.05{\tiny$\pm$0.30}                  & 34.37{\tiny$\pm$1.86}                  & 11.45{\tiny$\pm$0.38}                 & 90.87{\tiny$\pm$0.08}                  & 84.85{\tiny$\pm$0.11}                  & 19.48{\tiny$\pm$0.47}                  & 6.09{\tiny$\pm$0.12}                  \\
		\rowcolor{gray!10}BiomedCLIP (HT) & 4.1 M                            & \textcolor{red}{89.42{\tiny$\pm$0.54}} & \textcolor{red}{81.27{\tiny$\pm$0.84}} & \textcolor{red}{15.94{\tiny$\pm$0.47}} & \textcolor{red}{6.15{\tiny$\pm$0.25}} & \textcolor{red}{79.77{\tiny$\pm$0.55}} & \textcolor{red}{69.74{\tiny$\pm$0.65}} & \textcolor{red}{20.57{\tiny$\pm$0.75}} & \textcolor{red}{7.04{\tiny$\pm$0.21}} & \textcolor{red}{91.78{\tiny$\pm$0.10}} & \textcolor{red}{86.75{\tiny$\pm$0.12}} & \textcolor{red}{12.62{\tiny$\pm$0.38}} & \textcolor{red}{4.23{\tiny$\pm$0.19}} \\
		\bottomrule
	\end{tabular}
	\label{tab:results_seg}
\end{table*}

%% file: tab/results_cls.tex
\begin{table*}[htb]
	\caption{Classification results on three ultrasound datasets, averaged over three runs with different random seeds. (HT) denotes hybrid-tuned models. Performance \textcolor{red}{improvements} over baselines are highlighted. Mean{\tiny$\pm$std} is reported.}
	\centering
	\scriptsize
	\setlength{\tabcolsep}{4pt}
	\begin{tabular}{c|lr|cc|cc|cc|cc}
		\toprule
		\multirow{2}{*}{}                                                                & \multirow{2}{*}{\textbf{Method}} & \multirow{2}{*}{\textbf{Params}} & \multicolumn{2}{c|}{\textbf{LN-INT}}   & \multicolumn{2}{c|}{\textbf{LN-EXT}}   & \multicolumn{2}{c|}{\textbf{BUSI}}     & \multicolumn{2}{c}{\textbf{Average}}                                                                                                                                                                       \\ \cmidrule{4-11}
		                                                                                 &                                  &                                  & \textbf{Acc}                           & \textbf{AUC}                           & \textbf{Acc}                           & \textbf{AUC}                           & \textbf{Acc}                           & \textbf{AUC}                           & \textbf{Acc}                           & \textbf{AUC}                           \\
		\midrule
		                                                                                 & CLIP                             & 3.0 M                            & 61.19{\tiny$\pm$0.00}                  & 58.39{\tiny$\pm$0.00}                  & 54.55{\tiny$\pm$0.00}                  & 71.75{\tiny$\pm$0.00}                  & 70.48{\tiny$\pm$0.00}                  & 69.96{\tiny$\pm$0.00}                  & 62.07{\tiny$\pm$0.00}                  & 66.70{\tiny$\pm$0.00}                  \\
		\rowcolor{gray!10}\cellcolor{white}                                              & CLIP (HT)                        & 4.4 M                            & 44.02{\tiny$\pm$1.61}                  & \textcolor{red}{63.78{\tiny$\pm$3.76}} & \textcolor{red}{62.48{\tiny$\pm$6.11}} & \textcolor{red}{78.48{\tiny$\pm$5.76}} & 35.50{\tiny$\pm$4.24}                  & 58.21{\tiny$\pm$4.50}                  & 47.33{\tiny$\pm$1.74}                  & \textcolor{red}{66.82{\tiny$\pm$4.51}} \\
		                                                                                 & MetaCLIP                         & 2.8 M                            & 57.19{\tiny$\pm$0.00}                  & 57.10{\tiny$\pm$0.00}                  & 56.68{\tiny$\pm$0.00}                  & 61.43{\tiny$\pm$0.00}                  & 62.60{\tiny$\pm$0.00}                  & 59.24{\tiny$\pm$0.00}                  & 58.82{\tiny$\pm$0.00}                  & 59.26{\tiny$\pm$0.00}                  \\
		\rowcolor{gray!10}\cellcolor{white}                                              & MetaCLIP (HT)                    & 4.1 M                            & 41.50{\tiny$\pm$3.38}                  & 47.62{\tiny$\pm$7.43}                  & 39.04{\tiny$\pm$9.11}                  & 37.06{\tiny$\pm$11.43}                 & 47.55{\tiny$\pm$5.32}                  & 53.54{\tiny$\pm$7.09}                  & 42.70{\tiny$\pm$5.37}                  & 46.07{\tiny$\pm$4.29}                  \\
		                                                                                 & UniMedCLIP                       & 2.8 M                            & 47.50{\tiny$\pm$10.09}                 & 47.83{\tiny$\pm$13.55}                 & 43.76{\tiny$\pm$5.28}                  & 37.34{\tiny$\pm$7.87}                  & 56.15{\tiny$\pm$12.63}                 & 53.04{\tiny$\pm$10.20}                 & 49.14{\tiny$\pm$6.72}                  & 46.07{\tiny$\pm$4.86}                  \\
		\rowcolor{gray!10}\cellcolor{white}                                              & UniMedCLIP (HT)                  & 4.1 M                            & 40.09{\tiny$\pm$2.90}                  & 45.14{\tiny$\pm$1.86}                  & \textcolor{red}{46.61{\tiny$\pm$2.01}} & \textcolor{red}{41.22{\tiny$\pm$6.07}} & 36.07{\tiny$\pm$5.98}                  & \textcolor{red}{56.22{\tiny$\pm$2.91}} & 40.92{\tiny$\pm$2.43}                  & \textcolor{red}{47.53{\tiny$\pm$2.98}} \\
		                                                                                 & BiomedCLIP                       & 2.8 M                            & 58.68{\tiny$\pm$0.00}                  & 61.42{\tiny$\pm$0.00}                  & 63.64{\tiny$\pm$0.00}                  & 64.75{\tiny$\pm$0.00}                  & 52.24{\tiny$\pm$0.00}                  & 70.59{\tiny$\pm$0.00}                  & 58.19{\tiny$\pm$0.00}                  & 65.59{\tiny$\pm$0.00}                  \\
		\rowcolor{gray!10}\cellcolor{white}\multirow{-8}{*}{\rotatebox{90}{Zero-shot}}   & BiomedCLIP (HT)                  & 4.1 M                            & 40.48{\tiny$\pm$0.36}                  & 56.74{\tiny$\pm$1.12}                  & 50.18{\tiny$\pm$1.47}                  & 50.22{\tiny$\pm$3.99}                  & 44.51{\tiny$\pm$4.10}                  & 63.30{\tiny$\pm$2.45}                  & 45.06{\tiny$\pm$1.51}                  & 56.75{\tiny$\pm$1.96}                  \\
		\midrule
		                                                                                 & ResNet18                         & 11.2 M                           & 87.40{\tiny$\pm$0.00}                  & 94.94{\tiny$\pm$0.54}                  & 74.69{\tiny$\pm$2.01}                  & 87.32{\tiny$\pm$1.08}                  & 90.62{\tiny$\pm$1.57}                  & 95.45{\tiny$\pm$0.33}                  & 84.24{\tiny$\pm$1.15}                  & 92.57{\tiny$\pm$0.64}                  \\
		                                                                                 & DINOv2                           & 7.7 K                            & 57.48{\tiny$\pm$0.00}                  & 66.82{\tiny$\pm$0.60}                  & 51.16{\tiny$\pm$0.41}                  & 69.72{\tiny$\pm$4.31}                  & 64.06{\tiny$\pm$2.70}                  & 64.27{\tiny$\pm$0.75}                  & 57.57{\tiny$\pm$0.94}                  & 66.94{\tiny$\pm$1.41}                  \\
		\cmidrule{2-11}
		                                                                                 & CLIP                             & 3.0 M                            & 86.61{\tiny$\pm$1.36}                  & 92.21{\tiny$\pm$0.47}                  & 74.78{\tiny$\pm$2.16}                  & 83.88{\tiny$\pm$3.67}                  & 82.29{\tiny$\pm$0.90}                  & 90.94{\tiny$\pm$1.81}                  & 81.23{\tiny$\pm$1.11}                  & 89.01{\tiny$\pm$1.04}                  \\
		\rowcolor{gray!10}\cellcolor{white}                                              & CLIP (HT)                        & 4.4 M                            & 86.61{\tiny$\pm$0.79}                  & \textcolor{red}{92.34{\tiny$\pm$1.35}} & 74.33{\tiny$\pm$4.45}                  & \textcolor{red}{87.21{\tiny$\pm$1.81}} & \textcolor{red}{82.81{\tiny$\pm$1.57}} & 90.17{\tiny$\pm$2.38}                  & \textcolor{red}{81.25{\tiny$\pm$1.04}} & \textcolor{red}{89.90{\tiny$\pm$1.29}} \\
		                                                                                 & MetaCLIP                         & 2.8 M                            & 84.25{\tiny$\pm$0.00}                  & 91.59{\tiny$\pm$0.47}                  & 74.69{\tiny$\pm$1.21}                  & 88.57{\tiny$\pm$0.41}                  & 82.29{\tiny$\pm$1.81}                  & 90.66{\tiny$\pm$1.32}                  & 80.41{\tiny$\pm$0.98}                  & 90.28{\tiny$\pm$0.46}                  \\
		\rowcolor{gray!10}\cellcolor{white}                                              & MetaCLIP (HT)                    & 4.1 M                            & \textcolor{red}{85.57{\tiny$\pm$2.76}} & \textcolor{red}{92.79{\tiny$\pm$1.29}} & \textcolor{red}{76.56{\tiny$\pm$4.14}} & 86.85{\tiny$\pm$2.66}                  & \textcolor{red}{82.81{\tiny$\pm$1.57}} & 85.52{\tiny$\pm$0.56}                  & \textcolor{red}{81.65{\tiny$\pm$0.12}} & 88.39{\tiny$\pm$0.97}                  \\
		                                                                                 & UniMedCLIP                       & 2.8 M                            & 83.46{\tiny$\pm$1.58}                  & 92.03{\tiny$\pm$0.92}                  & 75.13{\tiny$\pm$2.45}                  & 84.86{\tiny$\pm$1.36}                  & 84.38{\tiny$\pm$1.57}                  & 94.86{\tiny$\pm$0.86}                  & 80.99{\tiny$\pm$0.61}                  & 90.58{\tiny$\pm$0.52}                  \\
		\rowcolor{gray!10}\cellcolor{white}                                              & UniMedCLIP (HT)                  & 4.1 M                            & \textcolor{red}{85.56{\tiny$\pm$1.20}} & \textcolor{red}{93.59{\tiny$\pm$1.01}} & 74.51{\tiny$\pm$4.68}                  & \textcolor{red}{87.26{\tiny$\pm$2.16}} & \textcolor{red}{89.06{\tiny$\pm$0.00}} & 93.47{\tiny$\pm$1.71}                  & \textcolor{red}{83.04{\tiny$\pm$1.47}} & \textcolor{red}{91.44{\tiny$\pm$0.96}} \\
		                                                                                 & BiomedCLIP                       & 2.8 M                            & 84.52{\tiny$\pm$1.64}                  & 91.95{\tiny$\pm$0.76}                  & 65.42{\tiny$\pm$1.78}                  & 76.99{\tiny$\pm$3.02}                  & 81.25{\tiny$\pm$4.14}                  & 90.83{\tiny$\pm$2.09}                  & 77.06{\tiny$\pm$1.20}                  & 86.59{\tiny$\pm$1.43}                  \\
		\rowcolor{gray!10}\cellcolor{white}\multirow{-10}{*}{\rotatebox{90}{Supervised}} & BiomedCLIP (HT)                  & 4.1 M                            & \textcolor{red}{86.61{\tiny$\pm$0.79}} & \textcolor{red}{93.06{\tiny$\pm$1.37}} & \textcolor{red}{69.87{\tiny$\pm$1.88}} & \textcolor{red}{85.52{\tiny$\pm$0.71}} & \textcolor{red}{85.94{\tiny$\pm$1.56}} & \textcolor{red}{91.84{\tiny$\pm$0.21}} & \textcolor{red}{80.81{\tiny$\pm$0.88}} & \textcolor{red}{90.14{\tiny$\pm$0.36}} \\
		\bottomrule
	\end{tabular}
	\label{tab:results_cls}
\end{table*}

%% file: tab/results_few_shot.tex
\begin{table*}[htb]
	\caption{Few-Shot segmentation and classification results on all available datasets, averaged over three runs with different random seeds. LoRA and HT denote LoRA-tuned and hybrid-tuned models. Ratio represents the proportion of total data used for training. The \textcolor{red}{best} results are highlighted. Mean{\tiny$\pm$std} is reported.}
	\centering
	\scriptsize
	\setlength{\tabcolsep}{2.4pt}
	\begin{tabular}{c|l|llll|lllll}
		\toprule
		\multirow{2}{*}{\textbf{Ratio}}     & \multirow{2}{*}{\textbf{Method}} & \multicolumn{4}{c|}{\textbf{Segmentation Average}} & \multicolumn{5}{c}{\textbf{Classification Average}}                                                                                                                                                                                                                                                                                                    \\ \cmidrule{3-11}
		                                    &                                  & \textbf{Dice}                                      & \textbf{IoU}                                        & \textbf{HD95}                          & \textbf{ASD}                           & \textbf{Acc}                           & \textbf{Rec}                            & \textbf{Pre}                            & \textbf{F1}                             & \textbf{AUC}                            \\
		\midrule
		\multirow{3}{*}{1\%}                & BiomedCLIP                       & \textcolor{red}{63.98{\tiny$\pm$0.99}}             & 51.64{\tiny$\pm$1.00}                               & 41.18{\tiny$\pm$2.35}                  & 15.58{\tiny$\pm$0.56}                  & \textcolor{red}{60.60{\tiny$\pm$3.03}} & 21.46{\tiny$\pm$3.83}                   & 57.78{\tiny$\pm$6.36}                   & 30.74{\tiny$\pm$4.37}                   & 62.99{\tiny$\pm$11.05}                  \\
		                                    & LoRA                             & 58.01{\tiny$\pm$1.91}                              & 45.97{\tiny$\pm$1.60}                               & 41.08{\tiny$\pm$1.69}                  & 15.12{\tiny$\pm$0.63}                  & 56.14{\tiny$\pm$1.84}                  & 6.88{\tiny$\pm$6.03}                    & 63.64{\tiny$\pm$17.39}                  & 10.62{\tiny$\pm$8.01}                   & 64.77{\tiny$\pm$13.48}                  \\
		\rowcolor{gray!10}\cellcolor{white} & HT                               & 63.55{\tiny$\pm$1.55}                              & \textcolor{red}{51.73{\tiny$\pm$1.71}}              & \textcolor{red}{39.05{\tiny$\pm$1.81}} & \textcolor{red}{14.64{\tiny$\pm$0.63}} & 59.26{\tiny$\pm$3.09}                  & \textcolor{red}{23.06{\tiny$\pm$3.32}}  & \textcolor{red}{68.55{\tiny$\pm$14.41}} & \textcolor{red}{32.93{\tiny$\pm$5.84}}  & \textcolor{red}{65.27{\tiny$\pm$11.51}} \\
		\cmidrule{1-11}
		\multirow{3}{*}{2\%}                & BiomedCLIP                       & 67.37{\tiny$\pm$0.95}                              & 55.30{\tiny$\pm$0.80}                               & 38.02{\tiny$\pm$1.47}                  & 14.31{\tiny$\pm$0.36}                  & \textcolor{red}{67.85{\tiny$\pm$7.46}} & 35.22{\tiny$\pm$12.37}                  & \textcolor{red}{75.64{\tiny$\pm$20.22}} & 44.95{\tiny$\pm$15.28}                  & 68.80{\tiny$\pm$11.33}                  \\
		                                    & LoRA                             & 62.28{\tiny$\pm$2.60}                              & 50.06{\tiny$\pm$1.97}                               & 41.51{\tiny$\pm$3.19}                  & 15.35{\tiny$\pm$1.90}                  & 65.44{\tiny$\pm$7.58}                  & \textcolor{red}{37.63{\tiny$\pm$10.74}} & 67.19{\tiny$\pm$18.94}                  & \textcolor{red}{46.16{\tiny$\pm$16.35}} & 69.52{\tiny$\pm$13.54}                  \\
		\rowcolor{gray!10}\cellcolor{white} & HT                               & \textcolor{red}{68.36{\tiny$\pm$1.29}}             & \textcolor{red}{56.91{\tiny$\pm$1.26}}              & \textcolor{red}{34.50{\tiny$\pm$1.28}} & \textcolor{red}{12.79{\tiny$\pm$0.45}} & 67.45{\tiny$\pm$8.94}                  & 31.89{\tiny$\pm$21.41}                  & 71.94{\tiny$\pm$17.24}                  & 42.07{\tiny$\pm$25.59}                  & \textcolor{red}{70.93{\tiny$\pm$10.26}} \\
		\cmidrule{1-11}
		\multirow{3}{*}{5\%}                & BiomedCLIP                       & 69.82{\tiny$\pm$0.66}                              & 58.38{\tiny$\pm$0.47}                               & 34.06{\tiny$\pm$0.97}                  & 12.54{\tiny$\pm$0.34}                  & \textcolor{red}{68.35{\tiny$\pm$7.29}} & 46.16{\tiny$\pm$7.60}                   & \textcolor{red}{73.80{\tiny$\pm$10.23}} & 56.11{\tiny$\pm$9.28}                   & 71.14{\tiny$\pm$7.64}                   \\
		                                    & LoRA                             & 65.98{\tiny$\pm$0.84}                              & 53.41{\tiny$\pm$0.72}                               & 40.29{\tiny$\pm$2.24}                  & 14.85{\tiny$\pm$0.96}                  & 66.34{\tiny$\pm$5.64}                  & \textcolor{red}{61.12{\tiny$\pm$20.49}} & 72.13{\tiny$\pm$21.26}                  & \textcolor{red}{58.37{\tiny$\pm$6.73}}  & \textcolor{red}{74.39{\tiny$\pm$7.17}}  \\
		\rowcolor{gray!10}\cellcolor{white} & HT                               & \textcolor{red}{71.73{\tiny$\pm$0.25}}             & \textcolor{red}{60.81{\tiny$\pm$0.20}}              & \textcolor{red}{29.51{\tiny$\pm$0.60}} & \textcolor{red}{10.75{\tiny$\pm$0.33}} & 65.09{\tiny$\pm$7.69}                  & 35.09{\tiny$\pm$31.66}                  & 46.71{\tiny$\pm$40.45}                  & 39.52{\tiny$\pm$34.74}                  & 70.92{\tiny$\pm$5.95}                   \\
		\cmidrule{1-11}
		\multirow{3}{*}{10\%}               & BiomedCLIP                       & 72.37{\tiny$\pm$0.21}                              & 61.17{\tiny$\pm$0.21}                               & 33.12{\tiny$\pm$0.72}                  & 11.85{\tiny$\pm$0.51}                  & 73.09{\tiny$\pm$1.38}                  & \textcolor{red}{58.67{\tiny$\pm$9.62}}  & 75.65{\tiny$\pm$6.93}                   & \textcolor{red}{64.82{\tiny$\pm$4.93}}  & 78.90{\tiny$\pm$1.91}                   \\
		                                    & LoRA                             & 65.97{\tiny$\pm$1.95}                              & 53.14{\tiny$\pm$1.74}                               & 41.00{\tiny$\pm$2.09}                  & 15.32{\tiny$\pm$0.84}                  & 66.02{\tiny$\pm$2.19}                  & 25.61{\tiny$\pm$4.21}                   & \textcolor{red}{93.81{\tiny$\pm$4.41}}  & 34.89{\tiny$\pm$5.40}                   & \textcolor{red}{81.59{\tiny$\pm$1.38}}  \\
		\rowcolor{gray!10}\cellcolor{white} & HT                               & \textcolor{red}{74.68{\tiny$\pm$0.22}}             & \textcolor{red}{64.10{\tiny$\pm$0.24}}              & \textcolor{red}{27.30{\tiny$\pm$0.51}} & \textcolor{red}{9.81{\tiny$\pm$0.24}}  & \textcolor{red}{73.63{\tiny$\pm$2.42}} & 51.83{\tiny$\pm$7.56}                   & 80.98{\tiny$\pm$3.91}                   & 62.11{\tiny$\pm$5.84}                   & 80.38{\tiny$\pm$1.46}                   \\
		\cmidrule{1-11}
		\multirow{3}{*}{20\%}               & BiomedCLIP                       & 74.18{\tiny$\pm$0.23}                              & 63.34{\tiny$\pm$0.33}                               & 32.48{\tiny$\pm$0.76}                  & 11.42{\tiny$\pm$0.13}                  & 74.62{\tiny$\pm$0.44}                  & 54.03{\tiny$\pm$7.27}                   & 82.03{\tiny$\pm$7.62}                   & 63.84{\tiny$\pm$2.70}                   & 81.93{\tiny$\pm$3.39}                   \\
		                                    & LoRA                             & 66.94{\tiny$\pm$1.21}                              & 54.09{\tiny$\pm$1.21}                               & 43.54{\tiny$\pm$3.53}                  & 16.43{\tiny$\pm$1.49}                  & 70.35{\tiny$\pm$7.51}                  & 39.66{\tiny$\pm$21.48}                  & \textcolor{red}{85.13{\tiny$\pm$4.33}}  & 49.72{\tiny$\pm$19.39}                  & 82.66{\tiny$\pm$1.61}                   \\
		\rowcolor{gray!10}\cellcolor{white} & HT                               & \textcolor{red}{77.47{\tiny$\pm$0.09}}             & \textcolor{red}{67.48{\tiny$\pm$0.16}}              & \textcolor{red}{24.11{\tiny$\pm$0.53}} & \textcolor{red}{8.52{\tiny$\pm$0.19}}  & \textcolor{red}{76.12{\tiny$\pm$1.87}} & \textcolor{red}{58.49{\tiny$\pm$5.10}}  & 81.02{\tiny$\pm$3.14}                   & \textcolor{red}{67.18{\tiny$\pm$4.11}}  & \textcolor{red}{84.86{\tiny$\pm$1.57}}  \\
		\cmidrule{1-11}
		\multirow{3}{*}{35\%}               & BiomedCLIP                       & 75.42{\tiny$\pm$0.41}                              & 64.77{\tiny$\pm$0.54}                               & 30.68{\tiny$\pm$0.83}                  & 10.60{\tiny$\pm$0.19}                  & 75.69{\tiny$\pm$0.76}                  & 56.15{\tiny$\pm$4.19}                   & 83.05{\tiny$\pm$2.64}                   & 66.26{\tiny$\pm$2.42}                   & 83.93{\tiny$\pm$1.73}                   \\
		                                    & LoRA                             & 66.37{\tiny$\pm$0.64}                              & 53.59{\tiny$\pm$0.60}                               & 44.00{\tiny$\pm$3.67}                  & 16.73{\tiny$\pm$1.64}                  & 73.65{\tiny$\pm$0.97}                  & 56.20{\tiny$\pm$13.06}                  & 81.72{\tiny$\pm$8.52}                   & 62.12{\tiny$\pm$7.14}                   & 82.16{\tiny$\pm$1.34}                   \\
		\rowcolor{gray!10}\cellcolor{white} & HT                               & \textcolor{red}{78.94{\tiny$\pm$0.43}}             & \textcolor{red}{69.28{\tiny$\pm$0.52}}              & \textcolor{red}{22.24{\tiny$\pm$0.53}} & \textcolor{red}{7.83{\tiny$\pm$0.17}}  & \textcolor{red}{76.76{\tiny$\pm$3.60}} & \textcolor{red}{58.86{\tiny$\pm$10.79}} & \textcolor{red}{83.44{\tiny$\pm$1.41}}  & \textcolor{red}{67.83{\tiny$\pm$7.77}}  & \textcolor{red}{86.06{\tiny$\pm$1.57}}  \\
		\cmidrule{1-11}
		\multirow{3}{*}{50\%}               & BiomedCLIP                       & 76.55{\tiny$\pm$0.46}                              & 66.03{\tiny$\pm$0.59}                               & 29.48{\tiny$\pm$1.34}                  & 10.30{\tiny$\pm$0.40}                  & 75.69{\tiny$\pm$1.63}                  & 55.52{\tiny$\pm$1.35}                   & 83.47{\tiny$\pm$3.75}                   & 65.89{\tiny$\pm$1.87}                   & 82.71{\tiny$\pm$1.40}                   \\
		                                    & LoRA                             & 66.88{\tiny$\pm$0.59}                              & 54.01{\tiny$\pm$0.82}                               & 46.26{\tiny$\pm$2.86}                  & 17.42{\tiny$\pm$1.30}                  & 69.47{\tiny$\pm$6.22}                  & 33.15{\tiny$\pm$15.91}                  & \textcolor{red}{92.37{\tiny$\pm$3.04}}  & 42.85{\tiny$\pm$17.15}                  & 83.50{\tiny$\pm$1.97}                   \\
		\rowcolor{gray!10}\cellcolor{white} & HT                               & \textcolor{red}{80.27{\tiny$\pm$0.31}}             & \textcolor{red}{70.81{\tiny$\pm$0.37}}              & \textcolor{red}{21.61{\tiny$\pm$0.60}} & \textcolor{red}{7.54{\tiny$\pm$0.24}}  & \textcolor{red}{77.16{\tiny$\pm$1.53}} & \textcolor{red}{57.54{\tiny$\pm$5.00}}  & 86.60{\tiny$\pm$5.62}                   & \textcolor{red}{68.02{\tiny$\pm$2.90}}  & \textcolor{red}{86.51{\tiny$\pm$2.34}}  \\
		\bottomrule
	\end{tabular}
	\label{tab:results_few_shot}
\end{table*}

%% file: tab/cross_evaluation.tex
\begin{table}[htbp]
	\caption{Overview of cross-dataset generalization evaluation. The \textcolor{red}{best} results are highlighted.}
	\centering
	\scriptsize
	\setlength{\tabcolsep}{3pt}
	\begin{tabular}{l|l|l|lll>{\columncolor{gray!10}}l}
		\toprule
		                                                &                                       & Scope        & Baseline               & LoRA                                    & Mona                                    & HT                                      \\
		\midrule
		\multirow{6}{*}{\rotatebox{90}{Segmentation}}   & \multirow{3}{*}{\rotatebox{90}{Dice}} & In-domain    & 80.01\tiny{$\pm$8.44}  & 67.34\tiny{$\pm$11.86}                  & 83.87\tiny{$\pm$5.68}                   & \textcolor{red}{84.03\tiny{$\pm$5.62}}  \\
		                                                &                                       & Cross-domain & 51.36\tiny{$\pm$11.70} & 48.62\tiny{$\pm$12.74}                  & 54.66\tiny{$\pm$13.43}                  & \textcolor{red}{55.50\tiny{$\pm$13.34}} \\
		                                                &                                       & Overall      & 56.14\tiny{$\pm$15.50} & 51.74\tiny{$\pm$14.36}                  & 59.52\tiny{$\pm$16.58}                  & \textcolor{red}{60.25\tiny{$\pm$16.35}} \\ \cmidrule{2-7}
		                                                & \multirow{3}{*}{\rotatebox{90}{IoU}}  & In-domain    & 70.20\tiny{$\pm$10.54} & 54.57\tiny{$\pm$12.66}                  & 75.27\tiny{$\pm$7.54}                   & \textcolor{red}{75.47\tiny{$\pm$7.46}}  \\
		                                                &                                       & Cross-domain & 39.29\tiny{$\pm$10.86} & 36.43\tiny{$\pm$11.15}                  & 43.42\tiny{$\pm$12.76}                  & \textcolor{red}{44.17\tiny{$\pm$12.69}} \\
		                                                &                                       & Overall      & 44.44\tiny{$\pm$15.80} & 39.45\tiny{$\pm$13.22}                  & 48.73\tiny{$\pm$16.93}                  & \textcolor{red}{49.39\tiny{$\pm$16.74}} \\
		\midrule
		\multirow{6}{*}{\rotatebox{90}{Classification}} & \multirow{3}{*}{\rotatebox{90}{Acc}}  & In-domain    & 82.88\tiny{$\pm$3.33}  & 67.82\tiny{$\pm$8.78}                   & 84.57\tiny{$\pm$3.07}                   & \textcolor{red}{86.28\tiny{$\pm$1.16}}  \\
		                                                &                                       & Cross-domain & 58.90\tiny{$\pm$4.16}  & \textcolor{red}{61.89\tiny{$\pm$8.77}}  & 61.37\tiny{$\pm$4.72}                   & 60.99\tiny{$\pm$6.62}                   \\
		                                                &                                       & Overall      & 66.89\tiny{$\pm$12.24} & 63.87\tiny{$\pm$8.99}                   & 69.10\tiny{$\pm$11.99}                  & \textcolor{red}{69.42\tiny{$\pm$13.39}} \\ \cmidrule{2-7}
		                                                & \multirow{3}{*}{\rotatebox{90}{AUC}}  & In-domain    & 91.39\tiny{$\pm$1.53}  & 82.14\tiny{$\pm$1.76}                   & \textcolor{red}{93.06\tiny{$\pm$1.13}}  & 92.45\tiny{$\pm$1.10}                   \\
		                                                &                                       & Cross-domain & 62.48\tiny{$\pm$10.01} & \textcolor{red}{69.10\tiny{$\pm$14.01}} & 65.59\tiny{$\pm$12.35}                  & 63.35\tiny{$\pm$14.00}                  \\
		                                                &                                       & Overall      & 72.12\tiny{$\pm$16.19} & 73.45\tiny{$\pm$12.96}                  & \textcolor{red}{74.74\tiny{$\pm$16.63}} & 73.05\tiny{$\pm$18.07}                  \\
		\bottomrule
	\end{tabular}
	\label{tab:cross_evaluation}
\end{table}

%% file: tab/abla_seg.tex
\begin{table}[htb]
	\caption{Ablation studies of different adapters for supervised segmentation, averaged over three runs with different random seeds. The \textcolor{red}{best} and \textcolor{ForestGreen}{second best} results are highlighted. Mean{\tiny$\pm$std} is reported.}
	\centering
	\scriptsize
	\setlength{\tabcolsep}{3pt}
	\begin{tabular}{c|l|llll}
		\toprule
		                                          & \textbf{Method} & \textbf{Dice}                                  & \textbf{IoU}                                   & \textbf{HD95}                                  & \textbf{ASD}                                  \\
		\midrule
		\multirow{5}{*}{\rotatebox{90}{LN-INT}}   & Baseline        & 71.27{\tiny$\pm$0.32}                          & 60.25{\tiny$\pm$0.89}                          & 34.41{\tiny$\pm$1.73}                          & 11.49{\tiny$\pm$0.65}                         \\
		                                          & LoRA (V+T)      & 56.67{\tiny$\pm$1.85}                          & 42.87{\tiny$\pm$1.80}                          & 55.69{\tiny$\pm$6.51}                          & 21.41{\tiny$\pm$2.66}                         \\
		                                          & LoRA            & 56.75{\tiny$\pm$2.35}                          & 43.01{\tiny$\pm$2.40}                          & 56.30{\tiny$\pm$9.29}                          & 22.10{\tiny$\pm$4.10}                         \\
		                                          & Mona            & \textcolor{ForestGreen}{79.05{\tiny$\pm$0.22}} & \textcolor{ForestGreen}{69.77{\tiny$\pm$0.16}} & \textcolor{red}{20.80{\tiny$\pm$0.70}}         & \textcolor{ForestGreen}{7.59{\tiny$\pm$0.51}} \\
		\rowcolor{gray!10}\cellcolor{white}       & HT              & \textcolor{red}{79.45{\tiny$\pm$0.24}}         & \textcolor{red}{70.31{\tiny$\pm$0.29}}         & \textcolor{ForestGreen}{20.81{\tiny$\pm$0.26}} & \textcolor{red}{7.40{\tiny$\pm$0.15}}         \\
		\midrule
		\multirow{5}{*}{\rotatebox{90}{LN-EXT}}   & Baseline        & 68.07{\tiny$\pm$0.19}                          & 55.95{\tiny$\pm$0.06}                          & 34.72{\tiny$\pm$0.92}                          & 11.63{\tiny$\pm$0.33}                         \\
		                                          & LoRA (V+T)      & 61.83{\tiny$\pm$0.52}                          & 48.38{\tiny$\pm$0.13}                          & 49.41{\tiny$\pm$5.08}                          & 17.80{\tiny$\pm$2.53}                         \\
		                                          & LoRA            & 61.51{\tiny$\pm$2.96}                          & 48.21{\tiny$\pm$3.36}                          & 49.94{\tiny$\pm$8.45}                          & 18.48{\tiny$\pm$4.12}                         \\
		                                          & Mona            & \textcolor{red}{74.56{\tiny$\pm$0.73}}         & \textcolor{ForestGreen}{63.73{\tiny$\pm$0.73}} & \textcolor{ForestGreen}{25.22{\tiny$\pm$0.61}} & \textcolor{red}{8.38{\tiny$\pm$0.16}}         \\
		\rowcolor{gray!10}\cellcolor{white}       & HT              & \textcolor{ForestGreen}{74.46{\tiny$\pm$0.98}} & \textcolor{red}{63.84{\tiny$\pm$0.94}}         & \textcolor{red}{25.16{\tiny$\pm$0.58}}         & \textcolor{ForestGreen}{8.46{\tiny$\pm$0.18}} \\
		\midrule
		\multirow{5}{*}{\rotatebox{90}{BUSI}}     & Baseline        & 76.57{\tiny$\pm$0.64}                          & 64.98{\tiny$\pm$0.72}                          & 29.35{\tiny$\pm$2.29}                          & 9.85{\tiny$\pm$0.52}                          \\
		                                          & LoRA (V+T)      & 62.34{\tiny$\pm$4.91}                          & 49.23{\tiny$\pm$5.10}                          & 46.03{\tiny$\pm$13.73}                         & 16.49{\tiny$\pm$6.20}                         \\
		                                          & LoRA            & 63.28{\tiny$\pm$2.09}                          & 49.68{\tiny$\pm$2.34}                          & 45.44{\tiny$\pm$7.44}                          & 15.89{\tiny$\pm$2.80}                         \\
		                                          & Mona            & \textcolor{ForestGreen}{79.40{\tiny$\pm$0.23}} & \textcolor{ForestGreen}{68.89{\tiny$\pm$0.36}} & \textcolor{red}{21.40{\tiny$\pm$0.99}}         & \textcolor{ForestGreen}{7.06{\tiny$\pm$0.19}} \\
		\rowcolor{gray!10}\cellcolor{white}       & HT              & \textcolor{red}{79.73{\tiny$\pm$0.66}}         & \textcolor{red}{69.29{\tiny$\pm$0.67}}         & \textcolor{ForestGreen}{21.78{\tiny$\pm$1.44}} & \textcolor{red}{7.05{\tiny$\pm$0.30}}         \\
		\midrule
		\multirow{5}{*}{\rotatebox{90}{DDTI}}     & Baseline        & 88.55{\tiny$\pm$0.12}                          & 79.85{\tiny$\pm$0.18}                          & 18.49{\tiny$\pm$0.47}                          & 6.90{\tiny$\pm$0.10}                          \\
		                                          & LoRA (V+T)      & 81.26{\tiny$\pm$0.72}                          & 69.58{\tiny$\pm$0.82}                          & 28.76{\tiny$\pm$0.36}                          & 11.06{\tiny$\pm$0.17}                         \\
		                                          & LoRA            & 81.40{\tiny$\pm$2.08}                          & 69.57{\tiny$\pm$2.66}                          & 28.04{\tiny$\pm$1.93}                          & 11.02{\tiny$\pm$0.48}                         \\
		                                          & Mona            & \textcolor{red}{89.50{\tiny$\pm$0.31}}         & \textcolor{red}{81.44{\tiny$\pm$0.47}}         & \textcolor{ForestGreen}{16.33{\tiny$\pm$0.69}} & \textcolor{ForestGreen}{6.20{\tiny$\pm$0.16}} \\
		\rowcolor{gray!10}\cellcolor{white}       & HT              & \textcolor{ForestGreen}{89.42{\tiny$\pm$0.54}} & \textcolor{ForestGreen}{81.27{\tiny$\pm$0.84}} & \textcolor{red}{15.94{\tiny$\pm$0.47}}         & \textcolor{red}{6.15{\tiny$\pm$0.25}}         \\
		\midrule
		\multirow{5}{*}{\rotatebox{90}{TN3K}}     & Baseline        & 72.76{\tiny$\pm$0.36}                          & 61.05{\tiny$\pm$0.30}                          & 34.37{\tiny$\pm$1.86}                          & 11.45{\tiny$\pm$0.38}                         \\
		                                          & LoRA (V+T)      & 57.71{\tiny$\pm$1.31}                          & 44.59{\tiny$\pm$1.21}                          & 67.05{\tiny$\pm$2.94}                          & 26.21{\tiny$\pm$1.25}                         \\
		                                          & LoRA            & 55.34{\tiny$\pm$5.12}                          & 42.32{\tiny$\pm$4.77}                          & 72.94{\tiny$\pm$5.82}                          & 29.17{\tiny$\pm$4.45}                         \\
		                                          & Mona            & \textcolor{red}{79.84{\tiny$\pm$0.24}}         & \textcolor{red}{69.78{\tiny$\pm$0.23}}         & \textcolor{red}{19.97{\tiny$\pm$0.38}}         & \textcolor{red}{6.81{\tiny$\pm$0.09}}         \\
		\rowcolor{gray!10}\cellcolor{white}       & HT              & \textcolor{ForestGreen}{79.77{\tiny$\pm$0.55}} & \textcolor{ForestGreen}{69.74{\tiny$\pm$0.65}} & \textcolor{ForestGreen}{20.57{\tiny$\pm$0.75}} & \textcolor{ForestGreen}{7.04{\tiny$\pm$0.21}} \\
		\midrule
		\multirow{5}{*}{\rotatebox{90}{Prostate}} & Baseline        & 90.87{\tiny$\pm$0.08}                          & 84.85{\tiny$\pm$0.11}                          & 19.48{\tiny$\pm$0.47}                          & 6.09{\tiny$\pm$0.12}                          \\
		                                          & LoRA (V+T)      & 80.05{\tiny$\pm$1.10}                          & 68.48{\tiny$\pm$1.17}                          & 41.76{\tiny$\pm$0.57}                          & 15.21{\tiny$\pm$0.17}                         \\
		                                          & LoRA            & 79.95{\tiny$\pm$0.57}                          & 68.29{\tiny$\pm$0.83}                          & 41.67{\tiny$\pm$1.37}                          & 15.50{\tiny$\pm$0.51}                         \\
		                                          & Mona            & \textcolor{ForestGreen}{91.56{\tiny$\pm$0.24}} & \textcolor{ForestGreen}{86.47{\tiny$\pm$0.31}} & \textcolor{ForestGreen}{12.65{\tiny$\pm$0.36}} & \textcolor{red}{4.21{\tiny$\pm$0.11}}         \\
		\rowcolor{gray!10}\cellcolor{white}       & HT              & \textcolor{red}{91.78{\tiny$\pm$0.10}}         & \textcolor{red}{86.75{\tiny$\pm$0.12}}         & \textcolor{red}{12.62{\tiny$\pm$0.38}}         & \textcolor{ForestGreen}{4.23{\tiny$\pm$0.19}} \\
		\midrule
		\multirow{5}{*}{\rotatebox{90}{Average}}  & Baseline        & 78.02{\tiny$\pm$0.12}                          & 67.82{\tiny$\pm$0.22}                          & 28.47{\tiny$\pm$0.77}                          & 9.57{\tiny$\pm$0.21}                          \\
		                                          & LoRA (V+T)      & 66.64{\tiny$\pm$0.34}                          & 53.85{\tiny$\pm$0.38}                          & 48.12{\tiny$\pm$4.06}                          & 18.03{\tiny$\pm$1.70}                         \\
		                                          & LoRA            & 66.37{\tiny$\pm$1.84}                          & 53.51{\tiny$\pm$1.84}                          & 49.06{\tiny$\pm$3.84}                          & 18.69{\tiny$\pm$2.17}                         \\
		                                          & Mona            & \textcolor{ForestGreen}{82.32{\tiny$\pm$0.10}} & \textcolor{ForestGreen}{73.35{\tiny$\pm$0.04}} & \textcolor{red}{19.40{\tiny$\pm$0.19}}         & \textcolor{red}{6.71{\tiny$\pm$0.08}}         \\
		\rowcolor{gray!10}\cellcolor{white}       & HT              & \textcolor{red}{82.44{\tiny$\pm$0.25}}         & \textcolor{red}{73.53{\tiny$\pm$0.32}}         & \textcolor{ForestGreen}{19.48{\tiny$\pm$0.15}} & \textcolor{ForestGreen}{6.72{\tiny$\pm$0.11}} \\
		\bottomrule
	\end{tabular}
	\label{tab:abla_seg}
\end{table}

%% file: tab/abla_cls.tex
\begin{table}[htb]
	\caption{Ablation studies of different adapters for supervised classification, averaged over three runs with different random seeds. The \textcolor{red}{best} and \textcolor{ForestGreen}{second best} results are highlighted. Mean{\tiny$\pm$std} is reported.}
	\centering
	\scriptsize
	\setlength{\tabcolsep}{3pt}
	\begin{tabular}{c|l|lllll}
		\toprule
		                                         & \textbf{Method} & \textbf{Acc}                                    & \textbf{Rec}                                   & \textbf{Pre}                                   & \textbf{F1}                                    & \textbf{AUC}                                   \\
		\midrule
		\multirow{5}{*}{\rotatebox{90}{LN-INT}}  & Baseline        & \textcolor{ForestGreen}{84.52{\tiny$\pm$1.64}}  & \textcolor{ForestGreen}{79.01{\tiny$\pm$4.28}} & 83.66{\tiny$\pm$0.85}                          & \textcolor{ForestGreen}{81.23{\tiny$\pm$2.46}} & 91.95{\tiny$\pm$0.76}                          \\
		                                         & LoRA (V+T)      & 63.78{\tiny$\pm$9.08}                           & 53.70{\tiny$\pm$38.26}                         & 72.60{\tiny$\pm$22.06}                         & 51.43{\tiny$\pm$18.00}                         & 79.29{\tiny$\pm$0.91}                          \\
		                                         & LoRA            & 62.73{\tiny$\pm$6.01}                           & 14.81{\tiny$\pm$15.16}                         & 75.06{\tiny$\pm$21.78}                         & 23.27{\tiny$\pm$21.73}                         & 81.30{\tiny$\pm$0.88}                          \\
		                                         & Mona            & 83.20{\tiny$\pm$0.91}                           & 72.84{\tiny$\pm$1.07}                          & \textcolor{ForestGreen}{85.58{\tiny$\pm$2.74}} & 78.67{\tiny$\pm$0.82}                          & \textcolor{red}{93.27{\tiny$\pm$0.76}}         \\
		\rowcolor{gray!10}\cellcolor{white}      & HT              & \textcolor{red}{86.61{\tiny$\pm$0.79}}          & \textcolor{red}{80.25{\tiny$\pm$2.14}}         & \textcolor{red}{87.33{\tiny$\pm$2.44}}         & \textcolor{red}{83.60{\tiny$\pm$0.88}}         & \textcolor{ForestGreen}{93.06{\tiny$\pm$1.37}} \\
		\midrule
		\multirow{5}{*}{\rotatebox{90}{LN-EXT}}  & Baseline        & 65.42{\tiny$\pm$1.78}                           & 37.86{\tiny$\pm$3.62}                          & 82.40{\tiny$\pm$3.76}                          & 51.80{\tiny$\pm$3.48}                          & 76.99{\tiny$\pm$3.02}                          \\
		                                         & LoRA (V+T)      & \textcolor{red}{72.73{\tiny$\pm$9.50}}          & \textcolor{red}{67.39{\tiny$\pm$35.24}}        & 84.19{\tiny$\pm$19.86}                         & \textcolor{red}{67.66{\tiny$\pm$19.47}}        & \textcolor{ForestGreen}{89.80{\tiny$\pm$2.17}} \\
		                                         & LoRA            & \textcolor{ForestGreen}{71.12{\tiny$\pm$13.31}} & 42.21{\tiny$\pm$27.89}                         & \textcolor{red}{98.48{\tiny$\pm$1.33}}         & 54.92{\tiny$\pm$30.50}                         & \textcolor{red}{91.02{\tiny$\pm$2.98}}         \\
		                                         & Mona            & 67.38{\tiny$\pm$2.09}                           & 38.95{\tiny$\pm$2.20}                          & 88.44{\tiny$\pm$7.05}                          & 54.02{\tiny$\pm$2.68}                          & 84.43{\tiny$\pm$1.69}                          \\
		\rowcolor{gray!10}\cellcolor{white}      & HT              & 69.87{\tiny$\pm$1.88}                           & \textcolor{ForestGreen}{44.38{\tiny$\pm$4.91}} & \textcolor{ForestGreen}{88.84{\tiny$\pm$1.26}} & \textcolor{ForestGreen}{59.07{\tiny$\pm$4.15}} & 85.52{\tiny$\pm$0.71}                          \\
		\midrule
		\multirow{5}{*}{\rotatebox{90}{BUSI}}    & Baseline        & \textcolor{ForestGreen}{81.25{\tiny$\pm$4.14}}  & 63.89{\tiny$\pm$6.36}                          & 82.07{\tiny$\pm$6.44}                          & 71.84{\tiny$\pm$6.45}                          & 90.83{\tiny$\pm$2.09}                          \\
		                                         & LoRA (V+T)      & 75.52{\tiny$\pm$3.61}                           & 52.78{\tiny$\pm$23.70}                         & 83.39{\tiny$\pm$19.93}                         & 60.14{\tiny$\pm$10.57}                         & 83.19{\tiny$\pm$2.96}                          \\
		                                         & LoRA            & 72.91{\tiny$\pm$8.89}                           & 27.78{\tiny$\pm$23.70}                         & \textcolor{red}{100.00{\tiny$\pm$0.00}}        & 40.04{\tiny$\pm$27.86}                         & 82.98{\tiny$\pm$2.21}                          \\
		                                         & Mona            & \textcolor{red}{85.94{\tiny$\pm$4.13}}          & \textcolor{red}{69.45{\tiny$\pm$8.67}}         & 90.71{\tiny$\pm$3.73}                          & \textcolor{red}{78.57{\tiny$\pm$6.91}}         & \textcolor{red}{92.85{\tiny$\pm$1.57}}         \\
		\rowcolor{gray!10}\cellcolor{white}      & HT              & \textcolor{red}{85.94{\tiny$\pm$1.56}}          & \textcolor{ForestGreen}{68.06{\tiny$\pm$6.36}} & \textcolor{ForestGreen}{92.62{\tiny$\pm$2.28}} & \textcolor{ForestGreen}{78.29{\tiny$\pm$3.42}} & \textcolor{ForestGreen}{91.84{\tiny$\pm$0.21}} \\
		\midrule
		\multirow{5}{*}{\rotatebox{90}{Average}} & Baseline        & 77.06{\tiny$\pm$1.20}                           & 60.25{\tiny$\pm$1.54}                          & 82.71{\tiny$\pm$3.20}                          & 68.29{\tiny$\pm$1.69}                          & 86.59{\tiny$\pm$1.43}                          \\
		                                         & LoRA (V+T)      & 70.68{\tiny$\pm$6.86}                           & 57.96{\tiny$\pm$31.93}                         & 80.06{\tiny$\pm$19.96}                         & 59.74{\tiny$\pm$15.78}                         & 84.10{\tiny$\pm$1.38}                          \\
		                                         & LoRA            & 68.92{\tiny$\pm$7.88}                           & 28.27{\tiny$\pm$18.26}                         & \textcolor{red}{91.18{\tiny$\pm$6.82}}         & 39.41{\tiny$\pm$22.76}                         & 85.10{\tiny$\pm$0.92}                          \\
		                                         & Mona            & \textcolor{ForestGreen}{78.84{\tiny$\pm$1.61}}  & \textcolor{ForestGreen}{60.41{\tiny$\pm$3.16}} & 88.25{\tiny$\pm$1.89}                          & \textcolor{ForestGreen}{70.42{\tiny$\pm$2.84}} & \textcolor{red}{90.18{\tiny$\pm$0.47}}         \\
		\rowcolor{gray!10}\cellcolor{white}      & HT              & \textcolor{red}{80.81{\tiny$\pm$0.88}}          & \textcolor{red}{64.23{\tiny$\pm$3.91}}         & \textcolor{ForestGreen}{89.60{\tiny$\pm$1.98}} & \textcolor{red}{73.65{\tiny$\pm$2.26}}         & \textcolor{ForestGreen}{90.14{\tiny$\pm$0.36}} \\
		\bottomrule
	\end{tabular}
	\label{tab:abla_cls}
\end{table}

%% file: tab/abla_ht_module.tex
\begin{table}[htb]
	\caption{Ablation studies of different adapter modules for supervised segmentation and classification, averaged over three runs with different random seeds. The \textcolor{red}{best} and \textcolor{ForestGreen}{second best} results are highlighted. Mean{\tiny$\pm$std} is reported.}
	\centering
	\scriptsize
	\begin{subtable}[t]{\linewidth}
		\centering
		\setlength{\tabcolsep}{3pt}
		\caption{Segmentation Task}
		\begin{tabular}{l|llll}
			\toprule
			\textbf{Method}             & \textbf{Dice}                                  & \textbf{IoU}                                   & \textbf{HD95}                                  & \textbf{ASD}                                  \\
			\midrule
			Mona                        & 82.32{\tiny$\pm$0.10}                          & 73.35{\tiny$\pm$0.04}                          & \textcolor{ForestGreen}{19.40{\tiny$\pm$0.19}} & \textcolor{ForestGreen}{6.71{\tiny$\pm$0.08}} \\
			HT (NE only)                & \textcolor{ForestGreen}{82.43{\tiny$\pm$0.19}} & \textcolor{red}{73.53{\tiny$\pm$0.25}}         & 19.42{\tiny$\pm$0.22}                          & \textcolor{ForestGreen}{6.71{\tiny$\pm$0.14}} \\
			HT (FF only)                & 82.32{\tiny$\pm$0.09}                          & \textcolor{ForestGreen}{73.37{\tiny$\pm$0.11}} & \textcolor{red}{19.33{\tiny$\pm$0.19}}         & \textcolor{red}{6.70{\tiny$\pm$0.06}}         \\
			\rowcolor{gray!10}HT (Full) & \textcolor{red}{82.44{\tiny$\pm$0.25}}         & \textcolor{red}{73.53{\tiny$\pm$0.32}}         & 19.48{\tiny$\pm$0.15}                          & 6.72{\tiny$\pm$0.11}                          \\
			\bottomrule
		\end{tabular}
	\end{subtable}

	\begin{subtable}[t]{\linewidth}
		\vspace{1em}
		\centering
		\setlength{\tabcolsep}{3pt}
		\caption{Classification Task}
		\begin{tabular}{l|lllll}
			\toprule
			\textbf{Method}             & \textbf{Acc}                                   & \textbf{Rec}                                   & \textbf{Pre}                                   & \textbf{F1}                                    & \textbf{AUC}                                   \\
			\midrule
			Mona                        & 78.84{\tiny$\pm$1.61}                          & 60.41{\tiny$\pm$3.16}                          & 88.25{\tiny$\pm$1.89}                          & 70.42{\tiny$\pm$2.84}                          & \textcolor{ForestGreen}{90.18{\tiny$\pm$0.47}} \\
			HT (NE only)                & \textcolor{ForestGreen}{80.38{\tiny$\pm$0.58}} & \textcolor{ForestGreen}{62.96{\tiny$\pm$1.86}} & \textcolor{red}{90.17{\tiny$\pm$1.11}}         & \textcolor{ForestGreen}{72.38{\tiny$\pm$0.89}} & 89.94{\tiny$\pm$0.46}                          \\
			HT (FF only)                & 79.51{\tiny$\pm$2.21}                          & 62.01{\tiny$\pm$4.78}                          & 88.50{\tiny$\pm$1.76}                          & 71.52{\tiny$\pm$3.73}                          & \textcolor{red}{90.52{\tiny$\pm$0.92}}         \\
			\rowcolor{gray!10}HT (Full) & \textcolor{red}{80.81{\tiny$\pm$0.88}}         & \textcolor{red}{64.23{\tiny$\pm$3.91}}         & \textcolor{ForestGreen}{89.60{\tiny$\pm$1.98}} & \textcolor{red}{73.65{\tiny$\pm$2.26}}         & 90.14{\tiny$\pm$0.36}                          \\
			\bottomrule
		\end{tabular}
	\end{subtable}
	\label{tab:abla_ht_module}
\end{table}

%% file: tab/efficiency.tex
\begin{table}[htb]
	\caption{Inference efficiency comparison on the segmentation task. All adapter-based methods use BiomedCLIP~\cite{biomedclip_zhang_2023} (ViT-B/16) backbone with 224$\times$224 input. GFLOPs and FPS are measured on a single NVIDIA 4090D GPU with batch size 1. Trainable params are shown in parentheses.}
	\centering
	\scriptsize
	\setlength{\tabcolsep}{3pt}
	\begin{tabular}{l|rrrr}
		\toprule
		\textbf{Method}                   & \textbf{Params (M)} & \textbf{GFLOPs} & \textbf{FPS} & \textbf{Latency (ms)} \\
		\midrule
		UNet~\cite{unet_ronneberger_2015} & 1.63 (1.63)         & 0.60            & 514.4        & 1.94                  \\
		\midrule
		Baseline                          & 198.66 (2.76)       & 14.77           & 291.7        & 3.43                  \\
		LoRA~\cite{lora_hu_2022}          & 199.55 (3.65)       & 19.79           & 141.7        & 7.06                  \\
		Mona~\cite{mona_yin_2024}         & 200.01 (4.10)       & 14.99           & 120.1        & 8.33                  \\
		\rowcolor{gray!10}HT (Ours)       & 200.02 (4.12)       & 14.99           & 93.2         & 10.73                 \\
		\bottomrule
	\end{tabular}
	\label{tab:efficiency}
\end{table}

%% file: sec/6_conclusion.tex
\section{Conclusion}

In this paper, we presented a novel HT strategy to effectively adapt general VLFMs for medical ultrasound analysis. By freezing the pre-trained backbone and introducing a lightweight adapter equipped with frequency filtering and noise estimation modules, HT explicitly addresses the unique acoustic artifacts of ultrasound without disrupting learned semantic priors. Extensive experiments across six diverse datasets demonstrate that HT significantly outperforms standard PEFT methods and domain-specific VLFMs in both segmentation and classification tasks. Furthermore, HT exhibits remarkable data efficiency in few-shot scenarios and robust generalization under severe cross-domain shifts. While challenges remain in zero-shot calibration due to prediction bias shifts, our findings establish HT as a highly effective and parameter-efficient paradigm. By successfully bridging the profound modality gap, this work paves a viable path toward deploying robust foundational intelligence in clinical ultrasound workflows.

%% file: sec/7_credits.tex
\section{Acknowledgement}
This work was supported by General Research Funds of the Research Grant Council of Hong Kong (Reference no. 15102222 and 15102524).

%% file: sec/8_appendix.tex
\appendix
\section{Ensembled Prompts for Zero-shot Classification}
\label{sec:appendix_prompts}

To enhance the zero-shot classification performance of Vision-Language Models on ultrasound images, we designed specialized ensembled prompts that incorporate domain-specific clinical knowledge. Instead of using generic templates, we constructed 10 diverse prompts for each class (benign and malignant) for both lymph node and breast lesion classification tasks. The complete list of prompts is provided below.

\subsection{Lymph Node Prompts}

\textbf{Benign Lymph Node:}
\begin{enumerate}
	\footnotesize
	\setlength\itemsep{0em}
	\item A benign lymph node with an oval shape and a preserved fatty hilum
	\item A benign lymph node with a long-to-short axis ratio greater than 2
	\item A benign lymph node showing a clear, echogenic central hilum
	\item A benign lymph node with a smooth, well-defined border
	\item A benign lymph node characterized by its regular, oval morphology and homogeneous echotexture
	\item A benign lymph node with a thin, uniform cortex surrounding a prominent hilum
	\item A benign lymph node appearing as a well-defined, hypoechoic oval structure with a bright central hilum
	\item A benign lymph node featuring a distinct fatty hilum and regular shape
	\item A benign lymph node with normal morphology, including a visible hilum and uniform cortex
	\item A benign lymph node that is distinctly elongated and maintains its central echogenic hilum
\end{enumerate}

\textbf{Malignant Lymph Node:}
\begin{enumerate}
	\footnotesize
	\setlength\itemsep{0em}
	\item A malignant lymph node with a round shape and an absent or effaced hilum
	\item A malignant lymph node with a long-to-short axis ratio less than 2
	\item A malignant lymph node with loss of the central fatty hilum
	\item A malignant lymph node with an irregular, spiculated, or blurred border
	\item A malignant lymph node containing internal microcalcifications
	\item A malignant lymph node showing internal cystic necrosis or liquefaction
	\item A malignant lymph node that is markedly hypoechoic and has a heterogeneous texture
	\item A malignant lymph node with eccentric cortical thickening
	\item A malignant lymph node appearing as a round, solid mass with indistinct margins
	\item A malignant lymph node characterized by a round shape and heterogeneous internal echoes
\end{enumerate}

\subsection{Breast Lesion Prompts}

\textbf{Benign Breast Nodule:}
\begin{enumerate}
	\footnotesize
	\setlength\itemsep{0em}
	\item A benign nodule with an oval shape and circumscribed margins
	\item A benign nodule with a parallel orientation, appearing wider-than-tall
	\item A benign nodule, simple cyst which is anechoic with posterior acoustic enhancement
	\item A benign nodule that is well-circumscribed and has a homogeneous echo pattern
	\item A benign nodule with a smooth border and an oval shape
	\item A benign nodule appearing as a solid, oval, and circumscribed mass
	\item A benign nodule with a gently lobulated but well-defined margin
	\item A benign nodule that is isoechoic and has a distinct, thin echogenic capsule
	\item A benign nodule with an oval shape, parallel orientation, and circumscribed margin
	\item A benign nodule with regular morphology and well-defined borders
\end{enumerate}

\textbf{Malignant Breast Nodule:}
\begin{enumerate}
	\footnotesize
	\setlength\itemsep{0em}
	\item A malignant nodule with an irregular shape and spiculated margins
	\item A malignant nodule with a non-parallel orientation, appearing taller-than-wide
	\item A malignant nodule causing posterior acoustic shadowing
	\item A malignant nodule with indistinct or angular margins
	\item A malignant nodule containing internal microcalcifications
	\item A malignant nodule that is markedly hypoechoic and has an irregular shape
	\item A malignant nodule with a heterogeneous echo pattern and ill-defined borders
	\item A malignant nodule with microlobulated margins
	\item A malignant nodule that is irregular in shape and demonstrates posterior shadowing
	\item A malignant nodule with suspicious morphology, including an irregular shape and non-circumscribed margins
\end{enumerate}